\documentclass[nohyperref]{article}

\usepackage{amsmath,amsfonts,bm}

\def\eqref#1{equation~\ref{#1}}

\def\1{\bm{1}}

\DeclareMathAlphabet{\mathsfit}{\encodingdefault}{\sfdefault}{m}{sl}
\SetMathAlphabet{\mathsfit}{bold}{\encodingdefault}{\sfdefault}{bx}{n}

\DeclareMathOperator*{\argmin}{arg\,min}

\usepackage{url}

\usepackage{graphicx}
\usepackage{tabularx}
\usepackage{multirow}
\usepackage{float}
\usepackage{subcaption}
\usepackage{enumitem}
\usepackage{sidecap}
\usepackage{placeins}
\usepackage{booktabs}
\newcolumntype{Y}{>{\centering\arraybackslash}X}

\usepackage{amsmath}

\usepackage[ruled, vlined]{algorithm2e}

\usepackage[textwidth=0.7in]{todonotes}

\newcommand{\steven}[2][]{}

\newcommand{\Amin}[2][]{}

\newcommand{\hamid}[2][]{}
\newcommand{\Hamid}[2][]{}

\usepackage{hyperref}

\usepackage[accepted]{icml2022}
\usepackage{amsmath}
\usepackage{amssymb}
\usepackage{mathtools}
\usepackage{amsthm}

\usepackage[capitalize,noabbrev]{cleveref}

\theoremstyle{plain}

\theoremstyle{definition}

\theoremstyle{remark}

\newcommand{\methodname}{\emph{PII}}
\newcommand{\augname}{ColorShift}
\newcommand{\augnameshort}{CS}

\newcommand{\squishlist}{
 \begin{list}{$\bullet$}
  { \setlength{\itemsep}{0pt}
     \setlength{\parsep}{3pt}
     \setlength{\topsep}{3pt}
     \setlength{\partopsep}{0pt}
     \setlength{\leftmargin}{1.5em}
     \setlength{\labelwidth}{1em}
     \setlength{\labelsep}{0.5em} } }
     
\newcommand{\squishend}{
  \end{list}  }

\begin{document}

\twocolumn[
\icmltitle{Plug-In Inversion: Model-Agnostic Inversion for Vision with Data Augmentations}



\icmlsetsymbol{equal}{*}

\begin{icmlauthorlist}
\icmlauthor{Amin Ghiasi}{equal,umd}
\icmlauthor{Hamid Kazemi}{equal,umd}
\icmlauthor{Steven Reich}{equal,umd}
\icmlauthor{Chen Zhu}{umd}
\icmlauthor{Micah Goldblum}{nyu}
\icmlauthor{Tom Goldstein}{umd}
\end{icmlauthorlist}

\icmlaffiliation{umd}{Department of Computer Science, University of Maryland, College Park, USA}
\icmlaffiliation{nyu}{New York University Center for Data Science, New York, USA}

\icmlcorrespondingauthor{Amin Ghiasi}{ghiasi@umd.edu}

\icmlkeywords{Machine Learning, ICML}

\vskip 0.3in
]



\printAffiliationsAndNotice{\icmlEqualContribution} 

\begin{abstract}
Existing techniques for model inversion typically rely on hard-to-tune regularizers, such as total variation or feature regularization, which must be individually calibrated for each network in order to produce adequate images. In this work, we introduce Plug-In Inversion, which relies on a simple set of augmentations and does not require excessive hyper-parameter tuning.  Under our proposed augmentation-based scheme, the same set of augmentation hyper-parameters can be used for inverting a wide range of image classification models, regardless of input dimensions or the architecture. We illustrate the practicality of our approach by inverting Vision Transformers (ViTs) and Multi-Layer Perceptrons (MLPs) trained on the ImageNet dataset, tasks which to the best of our knowledge have not been successfully accomplished by any previous works. 
\end{abstract}

\section{Introduction}

Model inversion is an important tool for visualizing and interpreting behaviors inside neural architectures, understanding what models have learned, and explaining model behaviors. In general, model inversion seeks inputs that either activate a feature in the network (\emph{feature visualization}) or yield a high output response for a particular class (\emph{class inversion}) \citep{olah2017feature}. 
Model inversion and visualization has been a cornerstone of conceptual studies that reveal how networks decompose images into semantic information
\citep{zeiler2014visualizing, dosovitskiy2016inverting}. 
Over time, inversion methods have shifted from solving conceptual problems to solving practical ones. 
Saliency maps, for example, are image-specific model visualizations that reveal the inputs that most strong influence a model's decisions \citep{simonyan2014deep}. 

Recent advances in network architecture pose major challenges for existing model inversion schemes.  Convolutional Neural Networks (CNN) have long been the de-facto approach for computer vision tasks, and are the focus of nearly all research in the model inversion field. Recently, other architectures have emerged that achieve results competitive with CNNs. These include Vision Transformers \citep[ViTs;][]{dosovitskiy2021image}, which are based on self-attention layers, and MLP-Mixer \citep{tolstikhin2021mlp} and ResMLP \citep{touvron2021resmlp}, which are based on Multi Layer Perceptron layers.   Unfortunately, most existing model inversion methods either cannot be applied to these architectures, or are known to fail.  For example, the feature regularizer used in DeepInversion \citep{yin2020dreaming} cannot be applied to ViTs or MLP-based models because they do not include Batch Normalization layers \citep{ioffe2015batch}.

In this work, we focus on class inversion, the goal of which is to find interpretable images that maximize the score a classification model assigns to a chosen label without knowledge about the model's training data. Class inversion has been used for a variety of tasks including model interpretation \citep{mordvintsev2015inceptionism}, image synthesis \citep{santurkar2019image}, and data-free knowledge transfer \citep{yin2020dreaming}. However, current inversion methods have several key drawbacks. The quality of generated images is often highly sensitive to the weights assigned to regularization terms, so these hyper-parameters need to be carefully calibrated for each individual network. In addition, methods requiring batch norm parameters are not applicable to emerging architectures.

To overcome these limitations, we present \emph{Plug-In Inversion} (\methodname), an augmentation-based approach to class inversion.
\methodname{} does not require any explicit regularization, which eliminates the need to tune regularizer-specific hyper-parameters for each model or image instance.
We show that \methodname{} is able to invert CNNs, ViTs, and MLP networks using the same architecture-agnostic method, and with the same architecture-agnostic hyper-parameters. 

\newpage
We summarize our contributions as follows:
\squishlist
    \item We provide a detailed analysis of various augmentations and how they affect the quality of images produced via class inversion.
    \item We introduce \emph{Plug-In Inversion} (\methodname), a new class inversion technique based on these augmentations, and compare it to existing techniques.
    \item We apply \methodname{} to dozens of different pre-trained models of varying architecture, justifying the claim that it can be `plugged in' to most networks without modification.
    \item In particular, we show that \methodname{} succeeds in inverting ViTs and large MLP-based architectures, which to our knowledge has not previously been accomplished.
    \item Finally, we explore the potential for combining \methodname{} with prior methods.
\squishend


\section{Background}\label{background}

\subsection{Class inversion}\label{plugin:general}
In the basic procedure for class inversion, we begin with a pre-trained model $f$ and chosen target class $y$. We randomly initialize (and optionally pre-process) an image $\mathbf{x}$ in the input space of $f$. We then perform gradient descent to solve the optimization problem $\hat{x} = \argmin_{\mathbf{x}} \mathcal{L}(f(\mathbf{x}), y)$ for a chosen objective function $\mathcal{L}$ to produce a class image $\hat{x}$. For very shallow networks and small datasets, letting $\mathcal{L}$ be cross-entropy or even the negative confidence assigned to the true class can produce recognizable images with minimal pre-processing \citep{fredrikson2015model}. Modern deep neural networks, however, cannot be inverted as easily.

\subsection{Regularization}
Most prior work on class inversion for deep networks has focused on carefully designing the objective function to produce quality images. This entails combining a divergence term (e.g. cross-entropy) with one or more regularization terms (\emph{image priors}) meant to guide the optimization towards an image with `natural' characteristics. \emph{DeepDream} \citep{mordvintsev2015inceptionism}, following work on feature inversion \citep{mahendran2015understanding}, uses two such terms: $\mathcal{R}_{\ell_2}(\mathbf{x}) = \| \mathbf{x} \|_2^2$, which penalizes the magnitude of the image vector, and total variation, defined as $\mathcal{R}_{TV}(\mathbf{x}) 
  = \sum_{\substack{\Delta_i\in\{0,1\} \\ \Delta_j\in\{0,1\}}} \left( \sum_{i, j} (x_{i+\Delta_i, j+\Delta_j} - x_{i, j})^2 \right)^\frac{1}{2}$, 
which penalizes sharp changes over small distances. 

\emph{DeepInversion} \citep{yin2020dreaming} uses both of these regularizers, along with the feature regularizer $\mathcal{R}_{feat}(\mathbf{x}) = \sum_k \left( \| \mu_k(\mathbf{x}) - \hat{\mu}_k \|_2 + \| \sigma_k^2(\mathbf{x}) - \hat{\sigma}_k^2 \|_2 \right)$, 
where $\mu_k, \sigma_k^2$ are the batch mean and variance of the features output by the $k$-th convolutional layer, and $\hat{\mu}_k, \hat{\sigma}_k^2$ are corresponding Batch Normalization statistics stored in the model \citep{ioffe2015batch}. Naturally, this method is only applicable to models that use Batch Normalization, which leaves out ViTs, MLPs, and even some CNNs. Furthermore, the optimal weights for each regularizer in the objective function vary wildly depending on architecture and training set, which presents a barrier to easily applying such methods to a wide array of networks.


\def \varfigcen{0.084\linewidth} 
\def \varfigcend{0.126\linewidth}
\begin{figure*}[htbp!]
    \centering
        \setlength\tabcolsep{1.5pt}
        \begin{tabularx}{\linewidth}{ccccccccc|c}
            \multicolumn{9}{c|}{w/ Centering } & w/o Centering  \\ 
            
            Init & Step 1 & Step 2 & Step 3 & Step 4 & Step 5 & Step 6 & Step 7 &  Final & Final \\

            \raisebox{0.25\totalheight}{\includegraphics[width=\varfigcen]{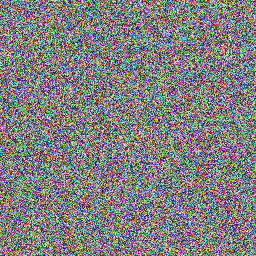}} &
            \raisebox{0.25\totalheight}{\includegraphics[width=\varfigcen]{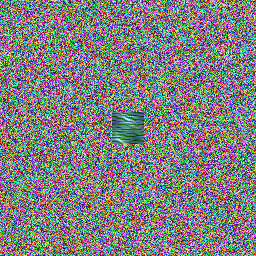}} &
            \raisebox{0.25\totalheight}{\includegraphics[width=\varfigcen]{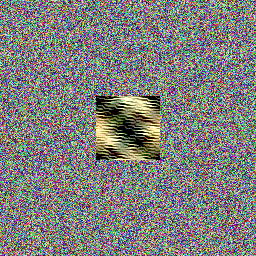}} &
            \raisebox{0.25\totalheight}{\includegraphics[width=\varfigcen]{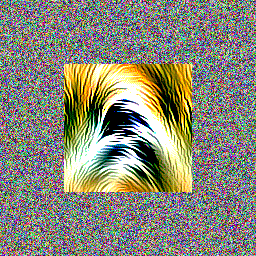}} &
            \raisebox{0.25\totalheight}{\includegraphics[width=\varfigcen]{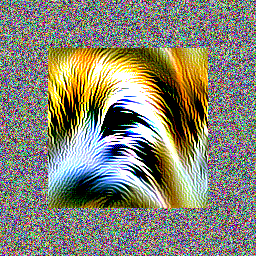}} &
            \raisebox{0.25\totalheight}{\includegraphics[width=\varfigcen]{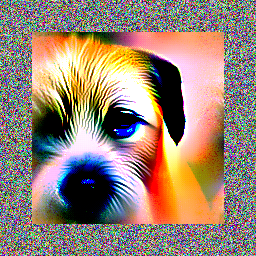}} &
            \raisebox{0.25\totalheight}{\includegraphics[width=\varfigcen]{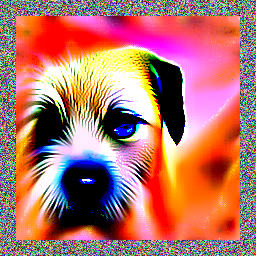}} &
            \raisebox{0.25\totalheight}{\includegraphics[width=\varfigcen]{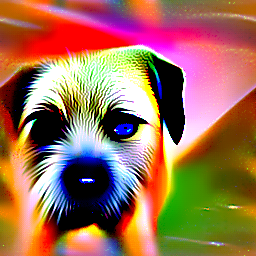}} &
            \includegraphics[width=\varfigcend]{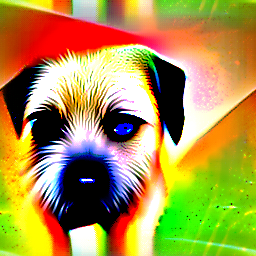} & 
            \includegraphics[width=\varfigcend]{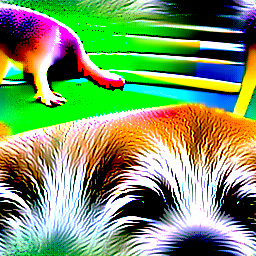} \\

        \end{tabularx}
        \caption{An image at different stages of optimization with centering (left), and an image inverted without centering (right), for the Border Terrier class of a robust ResNet-50.}
        \label{fig:centering}
\end{figure*}

\begin{figure*}[htbp!]
    \centering
        \centering
        \setlength\tabcolsep{1.5pt}
        \begin{tabularx}{\linewidth}{ccccccccc|c}
            \multicolumn{9}{c}{w/ Zoom} & w/o Zoom \\ 
            
            Init & Step 1 & Step 2 & Step 3 & Step 4 & Step 5 & Step 6 & Step 7 &  Final & Final \\

            \raisebox{0.25\totalheight}{\includegraphics[width=\varfigcen]{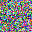}} &
            \raisebox{0.25\totalheight}{\includegraphics[width=\varfigcen]{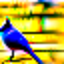}} &
            \raisebox{0.25\totalheight}{\includegraphics[width=\varfigcen]{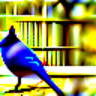}} &
            \raisebox{0.25\totalheight}{\includegraphics[width=\varfigcen]{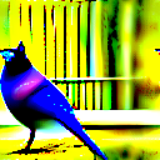}} &
            \raisebox{0.25\totalheight}{\includegraphics[width=\varfigcen]{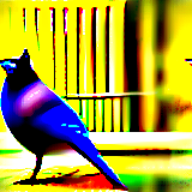}} &
            \raisebox{0.25\totalheight}{\includegraphics[width=\varfigcen]{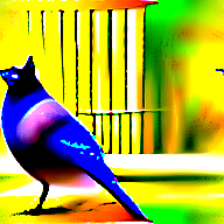}} &
            \raisebox{0.25\totalheight}{\includegraphics[width=\varfigcen]{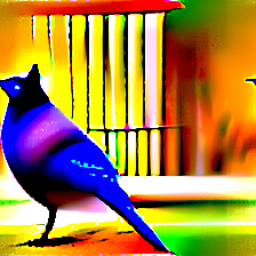}} &
            \raisebox{0.25\totalheight}{\includegraphics[width=\varfigcen]{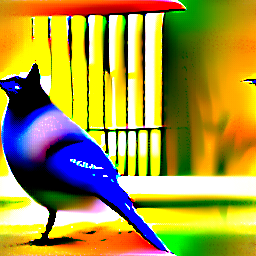}} &
            \includegraphics[width=\varfigcend]{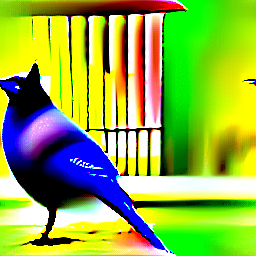} & 
            \includegraphics[width=\varfigcend]{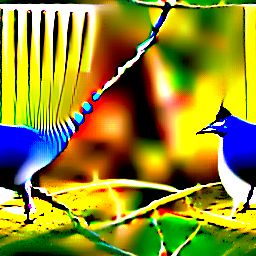} \\
          
        \end{tabularx}
        \caption{An image during different stages of optimization with zoom (left), and an image inverted without zoom (right), for the Jay class of a robust ResNet-50.}
        \label{fig:zoom}
\end{figure*}

\subsection{Architectures for vision}

We now present a brief overview of the three basic types of vision architectures that we will consider.

{\bf Convolutional Neural Networks} (CNNs) have long been the standard in deep learning for computer vision \citep{lecun1989backpropagation, krizhevsky2012imagenet}. Convolutional layers encourage a model to learn properties desirable for vision tasks, such as translation invariance. Numerous CNN models exist, mainly differing in the number, size, and arrangement of convolutional blocks and whether they include residual connections, Batch Normalization, or other modifications \citep{he2016deep, zagoruyko2016wide, simonyan2014very}.

\citet{dosovitskiy2021image} recently introduced {\bf Vision Transformers} (ViTs), adapting the Transformer architectures commonly used in NLP \citep{vaswani2017attention}. ViTs break input images into patches, combine them with positional embeddings, and use these as input tokens to self-attention modules. Some proposed variants require less training data \citep{touvron2021training}, have convolutional inductive biases \citep{d2021convit}, or make other modifications to the attention modules \citep{chu2021twins, liu2021swin, xu2021co}.

Subsequently, a number of authors have proposed vision models which are based solely on {\bf Multi-Layer Perceptrons} (MLPs), using insights from ViTs \citep{tolstikhin2021mlp, touvron2021resmlp, liu2021pay}. Generally, these models use patch embeddings similar to ViTs and alternate channel-wise and patch-wise linear embeddings, along with non-linearities and normalization.

We emphasize that as the latter two architecture types are recent developments, our work is the first to study them in the context of model inversion.

\section{Plug-In Inversion}\label{plugin}

Prior work on class inversion uses augmentations like jitter, which randomly shifts an image horizontally and vertically, and horizontal flips to improve the quality of inverted images
\citep{mordvintsev2015inceptionism, yin2020dreaming}. The hypothesis behind their use is that different views of the same image should result in similar scores for the target class. These augmentations are applied to the input before feeding it to the network, and different augmentations are used for each gradient step used to reconstruct $x$. In this section, we explore additional augmentations that benefit inversion before describing how we combine them to form the \methodname{} algorithm.

\begin{figure*}[h]
    \centering
    \setlength\tabcolsep{1.5pt}
    \begin{tabularx}{\linewidth}{cYYYYYY}
        $log(\lambda_{tv}):$ & $-9$ & $-8$ & $-7$ & $-6$ & $-5$ & $-4$ \\ 

        \raisebox{4\totalheight}{w/ \augnameshort} &
        \includegraphics[width=\linewidth]{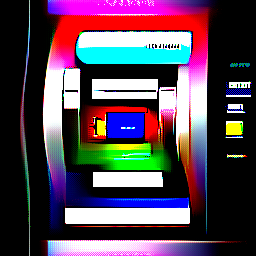}  &
        \includegraphics[width=\linewidth]{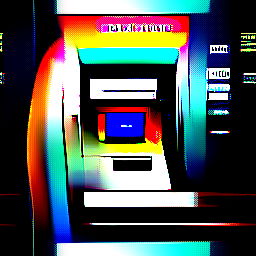}  &
        \includegraphics[width=\linewidth]{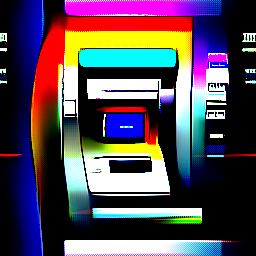}  &
        \includegraphics[width=\linewidth]{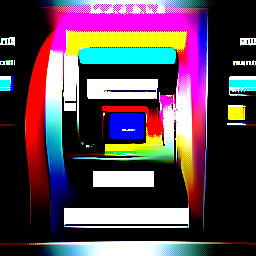}  &
        \includegraphics[width=\linewidth]{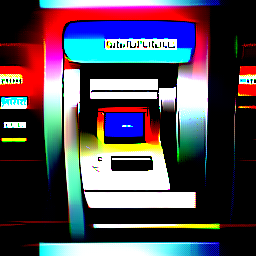}  &
        \includegraphics[width=\linewidth]{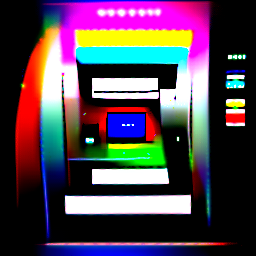}  \\   
        
        \raisebox{4\totalheight}{w/o \augnameshort} &
        \includegraphics[width=\linewidth]{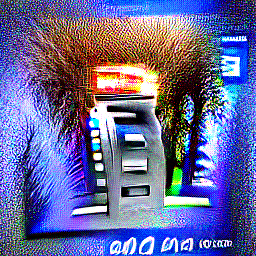}  &
        \includegraphics[width=\linewidth]{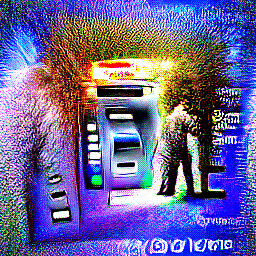}  &
        \includegraphics[width=\linewidth]{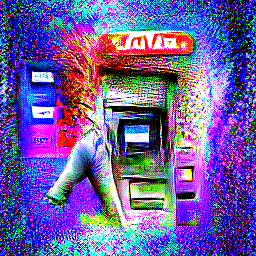}  &
        \includegraphics[width=\linewidth]{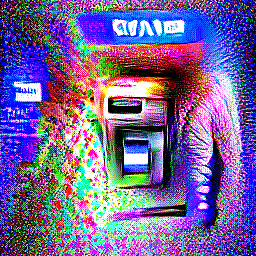}  &
        \includegraphics[width=\linewidth]{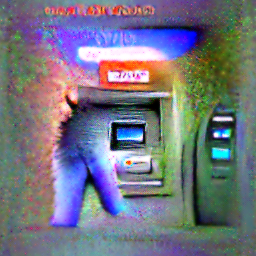}  &
        \includegraphics[width=\linewidth]{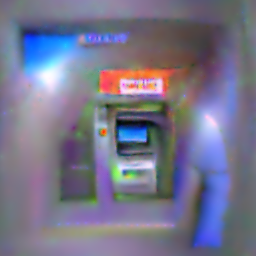}  \\

    \end{tabularx}
    \caption{ Inversions of the robust ResNet-50 ATM class, with and without ColorShift and with varying TV regularization strength. The inversion process with \augname{} is robust to changes in the $\lambda_{tv}$ hyper-parameter, while without it, $\lambda_{tv}$ seems to present a trade-off between noise and blur.}
    \label{fig:tv}
\end{figure*}

\begin{figure*}
    \centering
    \setlength\tabcolsep{1.5pt}
    \begin{tabularx}{\linewidth}{YYYYYYY}
            
        $e=1$ & $e=2$ & $e=4$ & $e=8$ & $e=16$ & $e=32$ & $e=64$ \\ 
             
        \includegraphics[width=\linewidth]{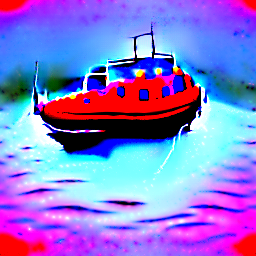} &
        \includegraphics[width=\linewidth]{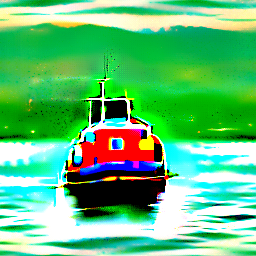} &
        \includegraphics[width=\linewidth]{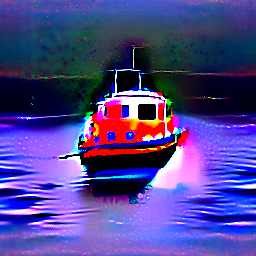} &
        \includegraphics[width=\linewidth]{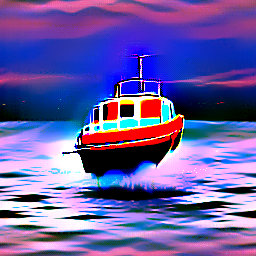} &
        \includegraphics[width=\linewidth]{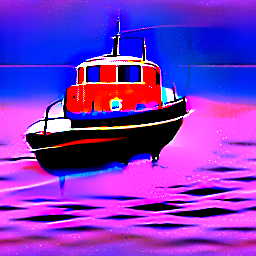} &
        \includegraphics[width=\linewidth]{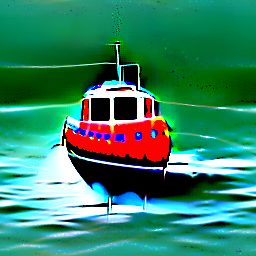} &
        \includegraphics[width=\linewidth]{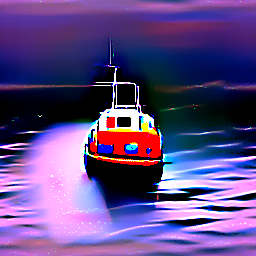} \\


            \end{tabularx}
        \caption{Effect of ensemble size in the quality of inverted images for the Tugboat class of a robust ResNet-50.}
        \label{fig:batch_size}
\end{figure*}

As robust models are typically easier to invert than naturally trained models \citep{santurkar2019image, mejia2019robust}, we use a robust ResNet-50 \citep{he2016deep} model trained on the ImageNet \citep{deng2009imagenet} dataset throughout this section as a toy example to examine how different augmentations impact inversion. Note, we perform the demonstrations in this section under slightly different conditions and with different models than those ultimately used for \methodname{} in order to highlight the effects of the augmentations as clearly as possible. The reader may find thorough experimental details in the appendix, section \ref{app:robust-setting}.

\subsection{Restricting Search Space}\label{plugin:search-space}

In this section, we consider two augmentations to improve the spatial qualities of inverted images:
\emph{Centering} and \emph{Zoom}. These are designed based on our hypothesis that restricting the input optimization space encourages better placement of recognizable features. Both methods start with small input patches, and each gradually increases this space in different ways to reach the intended input size. In doing so, they force the inversion algorithm to place important semantic content in the center of the image.

\paragraph{Centering}
Let $x$ be the input image being optimized. At first, we only optimize a patch at the center of $x$. After a fixed number of iterations, we increase the patch size outward by padding with random noise, repeating this until the patch reaches the full input size. Figure \ref{fig:centering} shows the state of the image prior at each stage of this process, as well as an image produced without centering. Without centering, the shift invariance of the networks allows most semantic content to scatter to the image edges.  With centering, results remain coherent. 

\paragraph{Zoom}
For zoom, we begin with an image $x$ of lower resolution than the desired result. In each step, we optimize this image for a fixed number of iterations and then up-sample the result, repeating until we reach the full resolution. Figure \ref{fig:zoom} shows the state of an image at each step of the zoom procedure, along with an image produced without zoom. The latter image splits the object of interest at its edges. By contrast, zoom appears to find a meaningful structure for the image in the early steps and refines details like texture as the resolution increases.

We note that zoom is not an entirely novel idea in inversion. \citet{yin2020dreaming} use a similar technique as `warm-up' for better performance and speed-up. However, we observe that continuing zoom throughout optimization contributes to the overall success of \methodname.


\def \varnatcen{0.088\linewidth}
\def \varnatcend{0.176\linewidth}
\begin{figure*}[h]
    \centering
        \centering
        \setlength\tabcolsep{1.5pt}
    \begin{tabularx}{\linewidth}{YYYY|YYYY}

           

             Z & Z + C & C & None & Z & Z + C & C & None \\  \includegraphics[width=\linewidth]{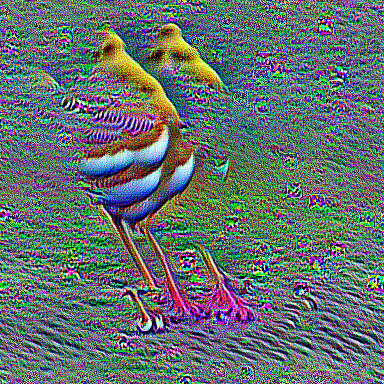} &
            \includegraphics[width=\linewidth]{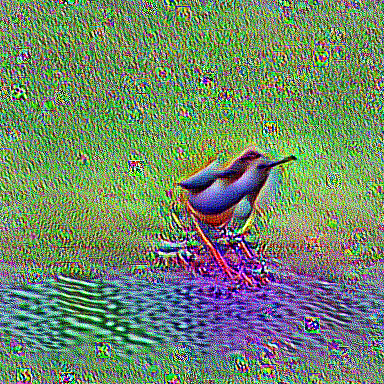} &
            \includegraphics[width=\linewidth]{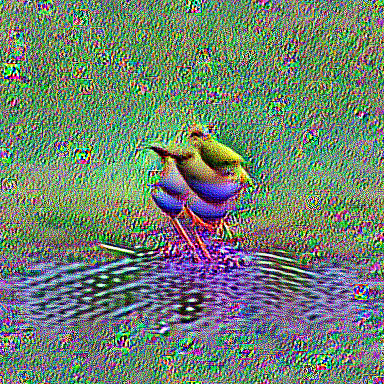} &
            \includegraphics[width=\linewidth]{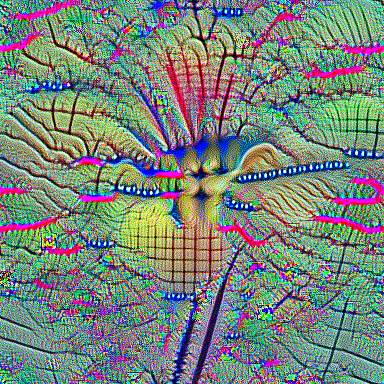} &
            \includegraphics[width=\linewidth]{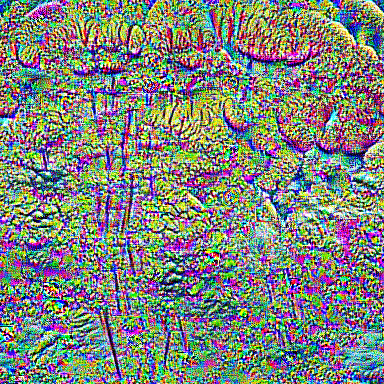} &
            \includegraphics[width=\linewidth]{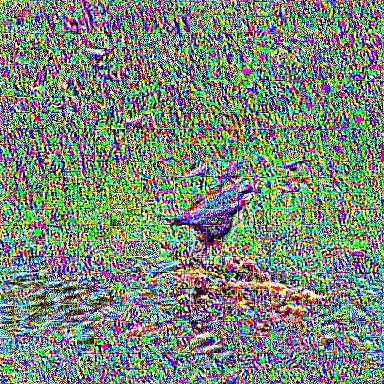} &
            \includegraphics[width=\linewidth]{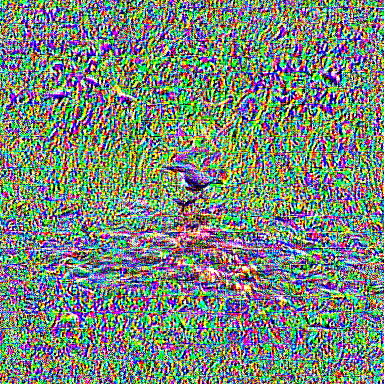} &
            \includegraphics[width=\linewidth]{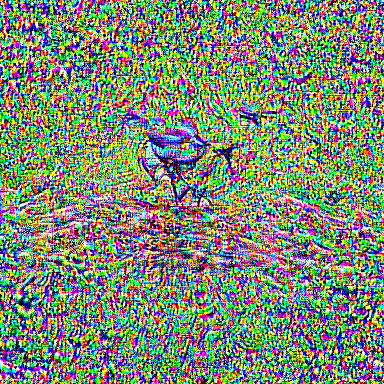}  \\
              \multicolumn{4}{c}{w/ ColorShift} & \multicolumn{4}{c}{w/o ColorShift}              \\
             
        \end{tabularx}
        \caption{The effect of various combinations of zoom, centering, and \augname{} when inverting the Dipper class using a \emph{naturally}-trained ResNet-50.}
        \label{fig:ablations}
\end{figure*}

\paragraph{Zoom + Centering}
Unsurprisingly, we have found that applying zoom and centering simultaneously yields even better results than applying either individually, since each one provides a different benefit. Centering places detailed and important features (e.g. the dog's eye in Figure \ref{fig:centering}) near the center and builds the rest of the image around the existing patch. Zoom helps enforce a sound large-scale structure for the image and fills in details later. 

The combined Zoom and Centering process proceeds in `stages', each at a higher resolution than the last.  Each stage begins with an image patch generated by the previous stage, which approximately minimizes the inversion loss. The patch is then up-sampled to a resolution halfway between the previous stage and current stage resolution, filling the center of the image and leaving a border which is padded with random noise. The next round of optimization then begins starting from this newly processed image.

\subsection{\augname{} Augmentation}\label{plugin:color-jitter}

The colors of the illustrative images we have shown so far are notably different from what one might expect in a natural image. This is due to \emph{ColorShift}, a new augmentation that we now present.

ColorShift is an adjustment of an image's colors by a random mean and variance in each channel. This can be formulated as follows:
\begin{equation*}
\text{\augname}(\mathbf{x}) = \sigma \mathbf{x} - \mu,
\end{equation*}
where $\mu$ and $\sigma$ are $C$-dimensional\footnote{$C$ being the number of channels} vectors drawn from $\mathcal{U}(-\alpha, \alpha)$ and $e^{\mathcal{U}(-\beta, \beta)}$, respectively, and are repeatedly redrawn after a fixed number of iterations. We use $\alpha = \beta = 1.0$ in all demonstrations unless otherwise noted.
At first glance, this deliberate shift away from the distribution of natural images seems counterproductive to the goal of producing a recognizable image. However, our results show that using ColorShift noticeably increases the amount of visual information in inverted images and also obviates the need for hard-to-tune regularizers to stabilize optimization. 

We visualize the stabilizing effect of ColorShift in Figure \ref{fig:tv}. In this experiment, we invert the model by minimizing the  sum of a cross entropy and a total-variation (TV) penalty.   Without ColorShift, the quality of images is highly dependent on the weight $\lambda_{TV}$ of the TV regularizer; smaller values produce noisy images, while larger values produce blurry ones. Inversion with ColorShift, on the other hand, is insensitive to this value and in fact succeeds when omitting the regularizer altogether.

Other preliminary experiments show that ColorShift similarly removes the need for $\ell_2$ or feature regularization, as our main results for \methodname{} will show. We conjecture that by forcing unnatural colors into an image, ColorShift requires the optimization to find a solution which contains meaningful semantic information, rather than photo-realistic colors, in order to achieve a high class score. Alternatively, as seen in Figure \ref{fig:distill_comp}, images optimized with an image prior may achieve high scores despite a lack of semantic information merely by finding sufficiently natural colors and textures.

\subsection{Ensembling}
Ensembling is an established tool often used in applications from enhanced inference~\citep{opitz1999popular} to dataset security~\citep{souri2021sleeper}. 
We find that optimizing an ensemble composed of different ColorShifts of the same image simultaneously improves the performance of inversion methods. 
To this end, we minimize the average of cross-entropy losses $\mathcal{L}(f(\mathbf{x}_i), y)$,
where the ${\mathbf{x}_i}$ are different ColorShifts of the image at the current step of optimization. Figure \ref{fig:batch_size} shows the result of applying ensembling alongside \augname{}. We observe that larger ensembles appear to give slight improvements, but even ensembles of size 1 or two produce satisfactory results. This is important for models like ViTs, where available GPU memory constrains the possible size of this ensemble; in general, we use the largest ensemble size (up to a maximum of $e=32$) that our hardware permits for a particular model. More results on the effect of ensemble size can be found in Figure \ref{fig:app:batch_size}. We show the effect of ensembling using other well-known augmentations and compare them to \augname{} in Appendix Section \ref{app:other_augs}, and observe that \augname{} is the strongest among augmentations we tried for model inversion.

\subsection{The Plug-in Inversion Method}\label{plugin:method}

We combine the jitter, ensembling, ColorShift, centering, and zoom techniques, and name the result Plug-In Inversion, which references the ability to `plug in' any differentiable model, including ViTs and MLPs, using a single fixed set of hyper-parameters. Full pseudocode for the algorithm may be found in appendix \ref{sec:algorithm}.
In the next section, we detail the experimental method that we used to find these hyper-parameters, after which we present our main results.

\def \varfiglen{0.195\linewidth} 
\begin{figure*}[h]
    \centering
        \centering
        \setlength\tabcolsep{0.5pt}
        \begin{tabularx}{\linewidth}{ccccc}
            \includegraphics[width=\varfiglen]{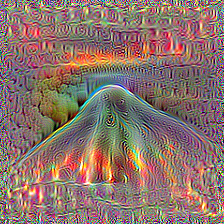} & 
            \includegraphics[width=\varfiglen]{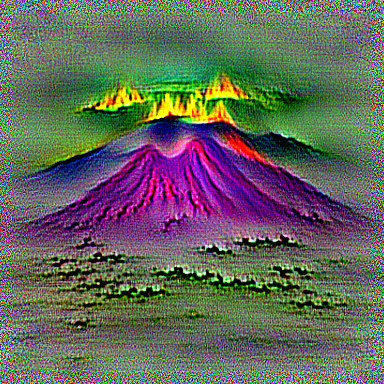} &
            \includegraphics[width=\varfiglen]{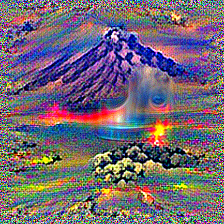} & 
            \includegraphics[width=\varfiglen]{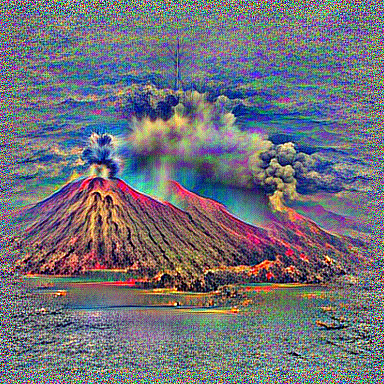} &
            \includegraphics[width=\varfiglen]{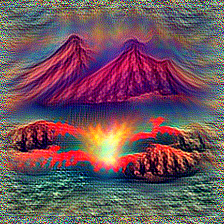} \\
            
            ResNet-101 & ViT B-32 & DeiT P16 224 & Deit Dist P16 384 & ConViT tiny \\ 
            
            \includegraphics[width=\varfiglen]{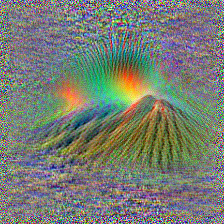} &
            \includegraphics[width=\varfiglen]{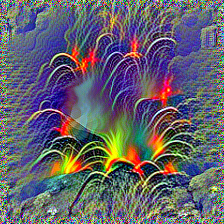} & 
            \includegraphics[width=\varfiglen]{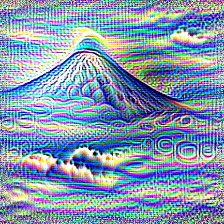} &
            \includegraphics[width=\varfiglen]{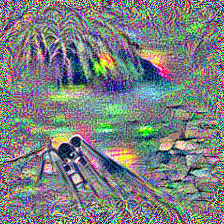} &
            \includegraphics[width=\varfiglen]{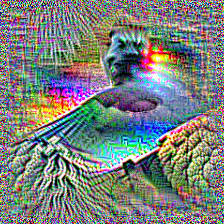} \\
            
            Mixer b16 224 & PiT Dist 224 & ResMLP 36 Dist & Swin P4 W12 & Twin PCPVT \\
            
        \end{tabularx}
        \caption{ Images inverted from the ImageNet Volcano class for various Convolutional, Transformer, and MLP-based networks using \methodname. 
        See figure \ref{fig:app_various_nets} for further examples. For more details about networks, refer to Appendix~\ref{app:model-library}. }
        \label{fig:various_nets}
\end{figure*}

\def \varvarnetvarclass{0.155\linewidth} 
\begin{figure*}[htbp!]
    \centering
        \centering
        \setlength\tabcolsep{1.5pt}
        \begin{tabularx}{\linewidth}{ccccccc}
            
            \multirow{2}{*}{} & \multirow{2}{*}{Barn} & Garbage & \multirow{2}{*}{Goblet} & Ocean & CRT & \multirow{2}{*}{Warplane}\\ 
              &  & Truck &  & Liner & Screen &  \\ 

\raisebox{0.3\totalheight}{\rotatebox[origin=lB]{90}{ResNet-101}} &
\includegraphics[width=\varvarnetvarclass]{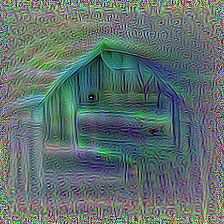}  &
\includegraphics[width=\varvarnetvarclass]{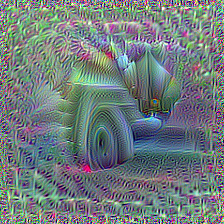}  &
\includegraphics[width=\varvarnetvarclass]{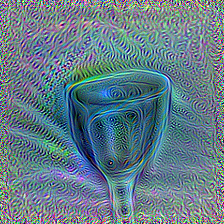}  &
\includegraphics[width=\varvarnetvarclass]{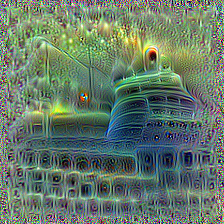}  &
\includegraphics[width=\varvarnetvarclass]{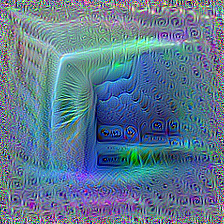}  &
\includegraphics[width=\varvarnetvarclass]{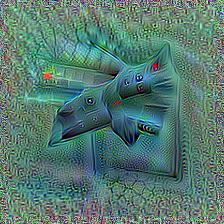}  \\

\raisebox{0.4\totalheight}{\rotatebox[origin=lB]{90}{ViT B-32}} &
\includegraphics[width=\varvarnetvarclass]{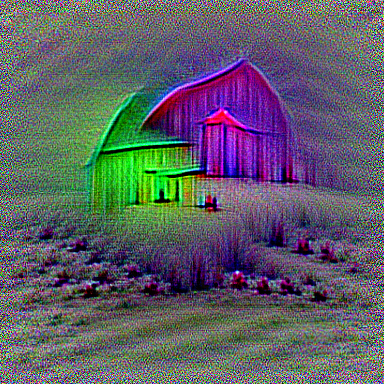}  &
\includegraphics[width=\varvarnetvarclass]{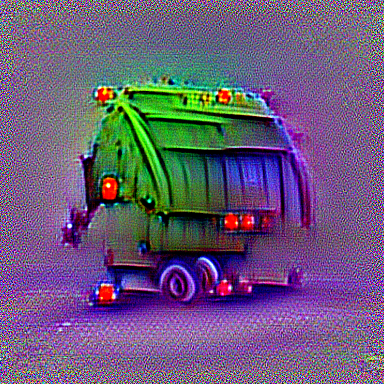}  &
\includegraphics[width=\varvarnetvarclass]{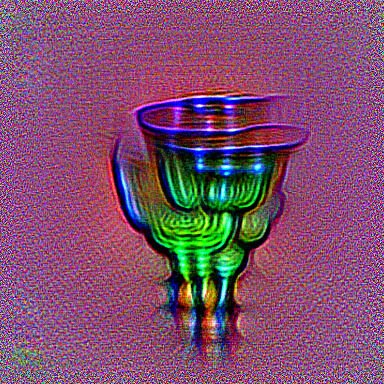}  &
\includegraphics[width=\varvarnetvarclass]{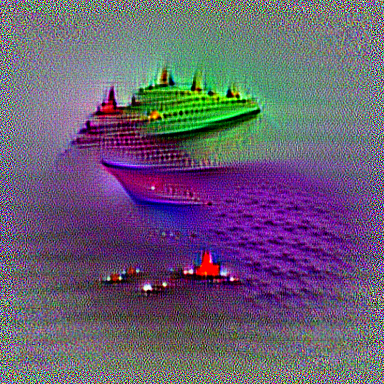}  &
\includegraphics[width=\varvarnetvarclass]{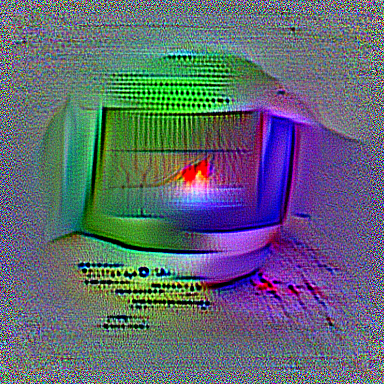}  &
\includegraphics[width=\varvarnetvarclass]{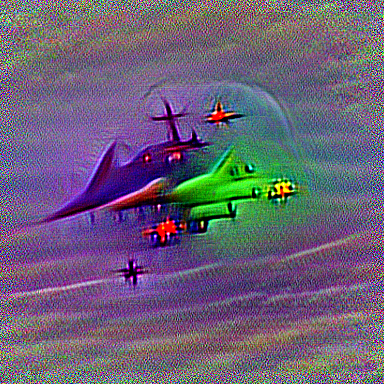}  \\

\raisebox{0.4\totalheight}{\rotatebox[origin=lB]{90}{DeiT Dist}} &
\includegraphics[width=\varvarnetvarclass]{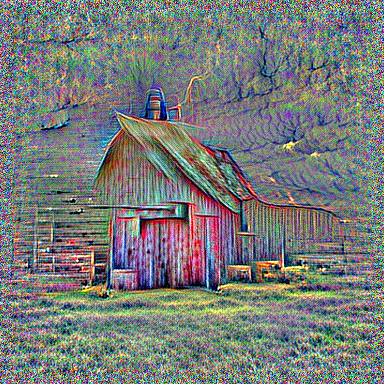}  &
\includegraphics[width=\varvarnetvarclass]{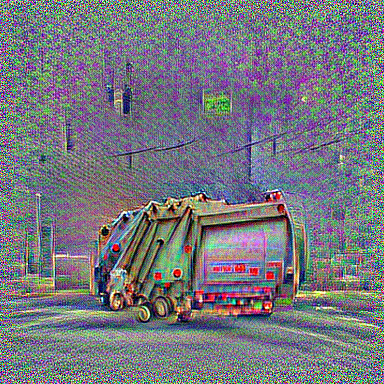}  &
\includegraphics[width=\varvarnetvarclass]{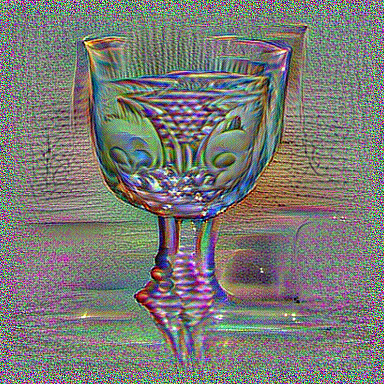}  &
\includegraphics[width=\varvarnetvarclass]{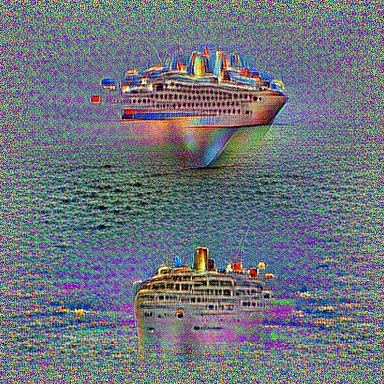}  &
\includegraphics[width=\varvarnetvarclass]{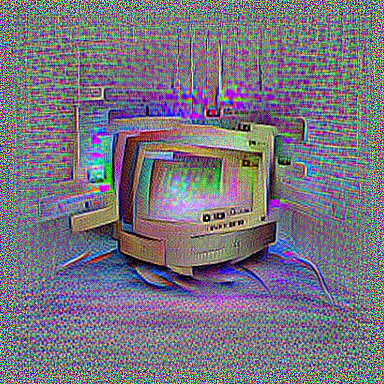}  &
\includegraphics[width=\varvarnetvarclass]{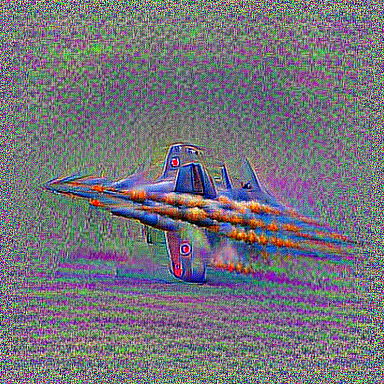}  \\

\raisebox{0.3\totalheight}{\rotatebox[origin=lB]{90}{ResMLP 36}} &
\includegraphics[width=\varvarnetvarclass]{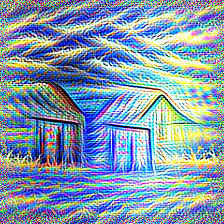}  &
\includegraphics[width=\varvarnetvarclass]{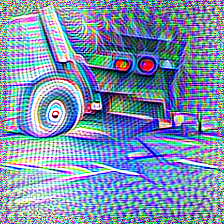}  &
\includegraphics[width=\varvarnetvarclass]{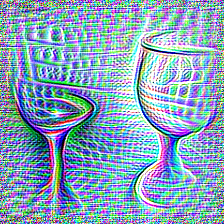}  &
\includegraphics[width=\varvarnetvarclass]{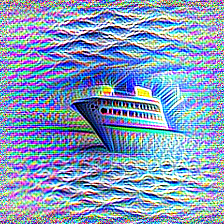}  &
\includegraphics[width=\varvarnetvarclass]{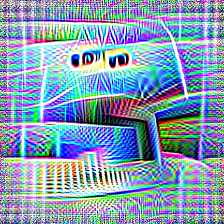}  &
\includegraphics[width=\varvarnetvarclass]{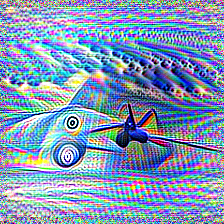}  \\

            \end{tabularx}
        \caption{Inverting different ImageNet model and class combinations for different classes using \methodname.}
        \label{fig:various_classes}
\end{figure*}


\def \cifar100{0.1023\linewidth} 
\begin{figure*}[h]
    \centering
        \centering
        \setlength\tabcolsep{1.5pt}
        \renewcommand{\arraystretch}{2}
        \begin{tabularx}{\linewidth}{ccccccccccc}
       & Apple & Castle & Dolphin & 
       Maple & Road & Rose & Sea & Seal & Train \\

\raisebox{0.1\totalheight}{\rotatebox[origin=lB]{90}{ViT L-16}} &
\includegraphics[width=\cifar100]{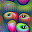}  &
\includegraphics[width=\cifar100]{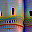}  &
\includegraphics[width=\cifar100]{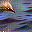}  &
\includegraphics[width=\cifar100]{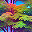}  &
\includegraphics[width=\cifar100]{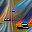}  &
\includegraphics[width=\cifar100]{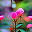}  &
\includegraphics[width=\cifar100]{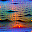}  &
\includegraphics[width=\cifar100]{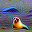}  &
\includegraphics[width=\cifar100]{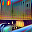}  \\

\raisebox{0.1\totalheight}{\rotatebox[origin=lB]{90}{ViT B-32}} &
\includegraphics[width=\cifar100]{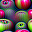}  &
\includegraphics[width=\cifar100]{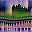}  &
\includegraphics[width=\cifar100]{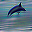}  &
\includegraphics[width=\cifar100]{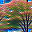}  &
\includegraphics[width=\cifar100]{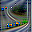}  &
\includegraphics[width=\cifar100]{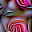}  &
\includegraphics[width=\cifar100]{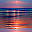}  &
\includegraphics[width=\cifar100]{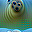}  &
\includegraphics[width=\cifar100]{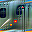}  \\

\raisebox{0.1\totalheight}{\rotatebox[origin=lB]{90}{ViT S-32}} &
\includegraphics[width=\cifar100]{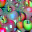}  &
\includegraphics[width=\cifar100]{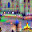}  &
\includegraphics[width=\cifar100]{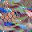}  &
\includegraphics[width=\cifar100]{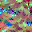}  &
\includegraphics[width=\cifar100]{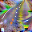}  &
\includegraphics[width=\cifar100]{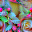}  &
\includegraphics[width=\cifar100]{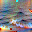}  &
\includegraphics[width=\cifar100]{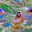}  &
\includegraphics[width=\cifar100]{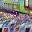}  \\

\raisebox{0.1\totalheight}{\rotatebox[origin=lB]{90}{ViT T-16}} &
\includegraphics[width=\cifar100]{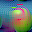}  &
\includegraphics[width=\cifar100]{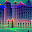}  &
\includegraphics[width=\cifar100]{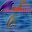}  &
\includegraphics[width=\cifar100]{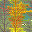}  &
\includegraphics[width=\cifar100]{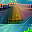}  &
\includegraphics[width=\cifar100]{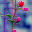}  &
\includegraphics[width=\cifar100]{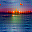}  &
\includegraphics[width=\cifar100]{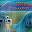}  &
\includegraphics[width=\cifar100]{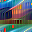}  \\

            \end{tabularx}
        \caption{Inverting different CIFAR-100 model and class combinations using \methodname.}
        \label{fig:cifar100}
\end{figure*}

\def \cifar10{0.0915\linewidth} 
\begin{figure*}[h]
    \centering
        \centering
        \setlength\tabcolsep{1.5pt}
        \renewcommand{\arraystretch}{2}
        \begin{tabularx}{\linewidth}{cccccccccccc}
       & Plane & Car & Bird & Cat &
       Deer & Dog & Frog & Horse & Ship & Truck \\
       
\raisebox{0.05\totalheight}{\rotatebox[origin=lB]{90}{ViT L-32}} &
\includegraphics[width=\cifar10]{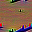}  &
\includegraphics[width=\cifar10]{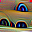}  &
\includegraphics[width=\cifar10]{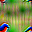}  &
\includegraphics[width=\cifar10]{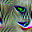}  &
\includegraphics[width=\cifar10]{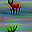}  &
\includegraphics[width=\cifar10]{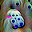}  &
\includegraphics[width=\cifar10]{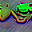}  &
\includegraphics[width=\cifar10]{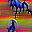}  &
\includegraphics[width=\cifar10]{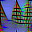}  &
\includegraphics[width=\cifar10]{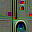}  \\

\raisebox{0.05\totalheight}{\rotatebox[origin=lB]{90}{ViT L-16}} &
\includegraphics[width=\cifar10]{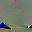}  &
\includegraphics[width=\cifar10]{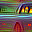}  &
\includegraphics[width=\cifar10]{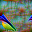}  &
\includegraphics[width=\cifar10]{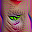}  &
\includegraphics[width=\cifar10]{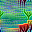}  &
\includegraphics[width=\cifar10]{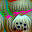}  &
\includegraphics[width=\cifar10]{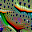}  &
\includegraphics[width=\cifar10]{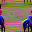}  &
\includegraphics[width=\cifar10]{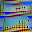}  &
\includegraphics[width=\cifar10]{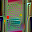}  \\

\raisebox{0.05\totalheight}{\rotatebox[origin=lB]{90}{ViT B-32}} &
\includegraphics[width=\cifar10]{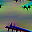}  &
\includegraphics[width=\cifar10]{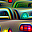}  &
\includegraphics[width=\cifar10]{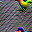}  &
\includegraphics[width=\cifar10]{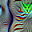}  &
\includegraphics[width=\cifar10]{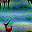}  &
\includegraphics[width=\cifar10]{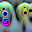}  &
\includegraphics[width=\cifar10]{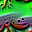}  &
\includegraphics[width=\cifar10]{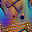}  &
\includegraphics[width=\cifar10]{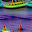}  &
\includegraphics[width=\cifar10]{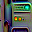}  \\

\raisebox{0.05\totalheight}{\rotatebox[origin=lB]{90}{ViT B-16}} &
\includegraphics[width=\cifar10]{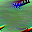}  &
\includegraphics[width=\cifar10]{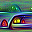}  &
\includegraphics[width=\cifar10]{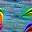}  &
\includegraphics[width=\cifar10]{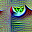}  &
\includegraphics[width=\cifar10]{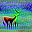}  &
\includegraphics[width=\cifar10]{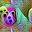}  &
\includegraphics[width=\cifar10]{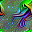}  &
\includegraphics[width=\cifar10]{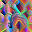}  &
\includegraphics[width=\cifar10]{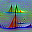}  &
\includegraphics[width=\cifar10]{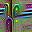}  \\

            \end{tabularx}
        \caption{Inverting different every class of CIFAR-10 from different ViT models using \methodname.}
        \label{fig:cifar10}
\end{figure*}

\section{Experimental Setup}\label{exp-setup}
In order to tune hyper-parameters of \methodname{} for use on naturally-trained models, we use the \texttt{torchvision} \citep{paszke2019pytorch} ImageNet-trained ResNet-50 model. We apply centering + zoom simultaneously in 7 `stages'. During each stage, we optimize the selected patch for 400 iterations, applying random jitter and \augname{} at each step. We use the Adam \citep{kingma2014adam} optimizer with momentum $\beta_m=(0.5, 0.99)$, initial learning rate $lr=0.01$, and cosine-decay. At the beginning of every stage, the learning rate and optimizer are re-initialized. We use $\alpha=\beta=1.0$ for the ColorShift parameters, and an ensemble size of $e=32$. Further ablation studies for these choices can be found in figures \ref{fig:app:mean_var}, \ref{fig:app:nat_centering}, and \ref{fig:app:batch_size}.

All the models (including pre-trained weights) we consider in this work are publicly available from widely-used sources. Explicit details of model resources can be found in section \ref{app:model-library} of the appendix. We also make the code used for all demonstrations and experiments in this work available at \url{https://github.com/youranonymousefriend/plugininversion}.


\section{Results}\label{results}

\subsection{\methodname{} works on a range of architectures}

We now present the results of applying Plug-In Inversion to different types of models. We once again emphasize that we use identical settings for the \methodname{} parameters in all cases.

Figure \ref{fig:various_nets} depicts images produced by inverting the Volcano class for a variety of architectures, including examples of CNNs, ViTs, and MLPs. While the quality of images varies somewhat between networks, all of them include distinguishable and well-placed visual information. Many more examples are found in Figure \ref{fig:app_various_nets} of the Appendix.

In Figure \ref{fig:various_classes}, we show images produced by \methodname{} from representatives of each main type of architecture for a few arbitrary ImageNet classes. We note the distinct visual styles that appear in each row, which supports the perspective of model inversion as a tool for understanding what kind of information different networks learn during training.

\def \vardistill{0.155\linewidth} 
\begin{figure*}[!h]
    \centering
        \centering
        \setlength\tabcolsep{1.5pt}
        \begin{tabularx}{\linewidth}{ccccccc}
            
            Gown & Microphone & Mobile Home & Schooner & Cardoon & Volcano \\ 
            
            
            \raisebox{2.5\totalheight}{\rotatebox[origin=lB]{90}{PII} } 
            \includegraphics[width=\vardistill]{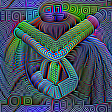} & 
            \includegraphics[width=\vardistill]{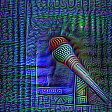} & 
            \includegraphics[width=\vardistill]{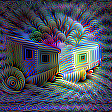} & 
            \includegraphics[width=\vardistill]{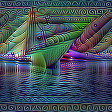} & 
            \includegraphics[width=\vardistill]{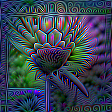} & 
            \includegraphics[width=\vardistill]{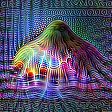} \\
            
            \raisebox{0.1\totalheight}{\rotatebox[origin=lB]{90}{PII + DeepInv} } 
            \includegraphics[width=\vardistill]{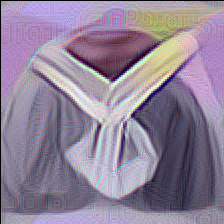} & 
            \includegraphics[width=\vardistill]{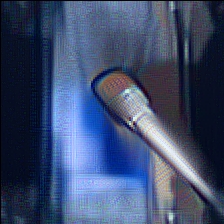} & 
            \includegraphics[width=\vardistill]{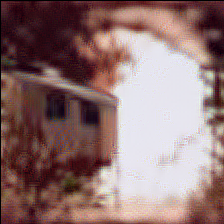} & 
            \includegraphics[width=\vardistill]{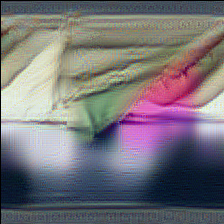} & 
            \includegraphics[width=\vardistill]{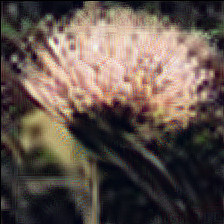} & 
            \includegraphics[width=\vardistill]{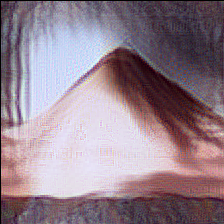} \\
            
            \raisebox{0.6\totalheight}{\rotatebox[origin=lB]{90}{DeepInv} } 
            \includegraphics[width=\vardistill]{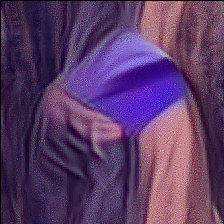} & 
            \includegraphics[width=\vardistill]{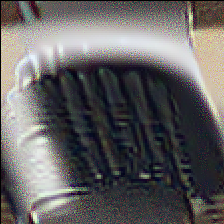} & 
            \includegraphics[width=\vardistill]{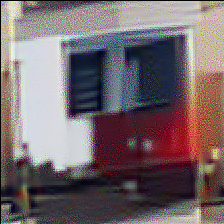} & 
            \includegraphics[width=\vardistill]{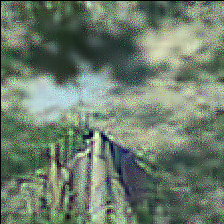} & 
            \includegraphics[width=\vardistill]{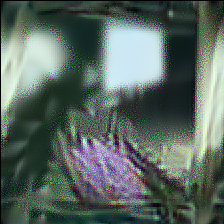} & 
            \includegraphics[width=\vardistill]{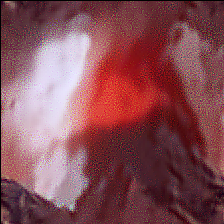} \\

        \end{tabularx}
        \caption{ \methodname{} and DeepInversion results for a naturally-trained ResNet-50. The middle row represents performing \methodname{} and using the result as an initialization for DeepInversion.}
        \label{fig:distill_comp}
\end{figure*}

\subsection{\methodname{} works on other datasets}

In Figure \ref{fig:cifar100}, we use \methodname{} to invert ViT models trained on ImageNet and fine-tuned on CIFAR-100. Figure \ref{fig:cifar10} shows inversion results from models fine-tuned on CIFAR-10. We emphasize that these were produced using identical settings to the ImageNet results above, whereas other methods (like DeepInversion) tune dataset-specific hyperparameters.

\subsection{Comparing \methodname{} to existing methods}

To quantitatively evaluate our method, we invert both a pre-trained ViT model and a pretrained ResMLP model to produce one image per class using \methodname{}, and do the same using DeepDream (i.e., DeepInversion minus feature regularization, which is not available for this model). We then use a variety of pre-trained models to classify these images. Table \ref{tab:pii_vs_dd} contains the mean top-1 and top-5 classification accuracies across these models, as well as Inception scores, for the generated images from each method. We see that our method is competitive with, and in the ViT case widely outperforms, DeepDream. Appendix \ref{sec:quant} contains more details about these experiments.

\begin{table}[h]
\setlength\tabcolsep{7pt}
\caption{Inception score and mean classification accuracies of various models on images inverted from (a) ViT B-32 and (b) ResMLP 36 by \methodname{} and DeepDream. Higher is better in all fields.}
    \begin{subtable}[h]{\columnwidth}
    \centering
    \begin{tabular}{c|ccc}
    \toprule
    Method      &   Inception score & Top-1     & Top-5     \\
    \midrule
    \methodname{}&  28.17 $\pm$ 7.21  & 77.0\%    & 89.5\%    \\
    DeepDream   &   2.72 $\pm$ 0.23   & 35.2\%    & 49.6\%    \\
    \bottomrule
    \end{tabular}
    \caption{Images inverted from ViT B-32}
    \end{subtable}
    \begin{subtable}[h]{\columnwidth}
    \centering
    \begin{tabular}{c|ccc}
    \toprule
    Method      &   Inception score & Top-1     & Top-5     \\
    \midrule
    \methodname{}&  $6.79 \pm 2.18$  & 49.2\%    & 62.0\%    \\
    DeepDream   &   $3.27 \pm 0.47$   & 51.3\%    & 61.3\%    \\
    \bottomrule
    \end{tabular}
    \caption{Images inverted from ResMLP 36}
    \end{subtable}
    \label{tab:pii_vs_dd}
\end{table}

Figure \ref{fig:distill_comp} shows images from a few arbitrary classes produced by \methodname{} and DeepInversion. We additionally show images produced by DeepInversion using the output of \methodname, rather than random noise, as its initialization. Using either initialization, DeepInversion clearly produces images with natural-looking colors and textures, which \methodname{} of course does not. However, DeepInversion alone results in some images that either do not clearly correspond to the target class or are semantically confusing. By comparison, \methodname{} again produces images with strong spatial and semantic qualities. Interestingly, these qualities appear to be largely retained when applying DeepInversion after \methodname, but with the color and texture improvements that image priors afford \citep{mahendran2015understanding}, suggesting that using these methods in tandem may be a way to produce even better inverted images from CNNs than either method independently.

Appendix \ref{sec:baselines} contains additional qualitative comparisons to DeepDream and DeepInversion, further illustrating the need for model-specific hyperparameter tuning in contrast to our method.

\section{Conclusion}

We studied the effect of various augmentations on the quality of class-inverted images and introduced Plug-In Inversion, which uses these augmentations in tandem. We showed that this technique produces intelligible images from a wide range of well-studied architectures and datasets, including the recently introduced ViTs and MLPs, without a need for model-specific hyper-parameter tuning. We believe that augmentation-based model inversion is a promising direction for future research in understanding computer vision models.



\clearpage
\bibliography{main}

\begin{thebibliography}{45}
\providecommand{\natexlab}[1]{#1}
\providecommand{\url}[1]{\texttt{#1}}
\expandafter\ifx\csname urlstyle\endcsname\relax
  \providecommand{\doi}[1]{doi: #1}\else
  \providecommand{\doi}{doi: \begingroup \urlstyle{rm}\Url}\fi

\bibitem[Chen et~al.(2020)Chen, Fan, Girshick, and He]{chen2020improved}
Chen, X., Fan, H., Girshick, R., and He, K.
\newblock Improved baselines with momentum contrastive learning.
\newblock \emph{arXiv preprint arXiv:2003.04297}, 2020.

\bibitem[Chu et~al.(2021)Chu, Tian, Wang, Zhang, Ren, Wei, Xia, and
  Shen]{chu2021twins}
Chu, X., Tian, Z., Wang, Y., Zhang, B., Ren, H., Wei, X., Xia, H., and Shen, C.
\newblock Twins: Revisiting the design of spatial attention in vision
  transformers.
\newblock \emph{arXiv preprint arXiv:2104.13840}, 1\penalty0 (2):\penalty0 3,
  2021.

\bibitem[d'Ascoli et~al.(2021)d'Ascoli, Touvron, Leavitt, Morcos, Biroli, and
  Sagun]{d2021convit}
d'Ascoli, S., Touvron, H., Leavitt, M., Morcos, A., Biroli, G., and Sagun, L.
\newblock Convit: Improving vision transformers with soft convolutional
  inductive biases.
\newblock \emph{arXiv preprint arXiv:2103.10697}, 2021.

\bibitem[Deng et~al.(2009)Deng, Dong, Socher, Li, Li, and
  Fei-Fei]{deng2009imagenet}
Deng, J., Dong, W., Socher, R., Li, L.-J., Li, K., and Fei-Fei, L.
\newblock Imagenet: A large-scale hierarchical image database.
\newblock In \emph{2009 IEEE conference on computer vision and pattern
  recognition}, pp.\  248--255. Ieee, 2009.

\bibitem[Dosovitskiy \& Brox(2016)Dosovitskiy and
  Brox]{dosovitskiy2016inverting}
Dosovitskiy, A. and Brox, T.
\newblock Inverting visual representations with convolutional networks.
\newblock In \emph{Proceedings of the IEEE conference on computer vision and
  pattern recognition}, pp.\  4829--4837, 2016.

\bibitem[Dosovitskiy et~al.(2021)Dosovitskiy, Beyer, Kolesnikov, Weissenborn,
  Zhai, Unterthiner, Dehghani, Minderer, Heigold, Gelly, Uszkoreit, and
  Houlsby]{dosovitskiy2021image}
Dosovitskiy, A., Beyer, L., Kolesnikov, A., Weissenborn, D., Zhai, X.,
  Unterthiner, T., Dehghani, M., Minderer, M., Heigold, G., Gelly, S.,
  Uszkoreit, J., and Houlsby, N.
\newblock An image is worth 16x16 words: Transformers for image recognition at
  scale, 2021.

\bibitem[Fredrikson et~al.(2015)Fredrikson, Jha, and
  Ristenpart]{fredrikson2015model}
Fredrikson, M., Jha, S., and Ristenpart, T.
\newblock Model inversion attacks that exploit confidence information and basic
  countermeasures.
\newblock In \emph{Proceedings of the 22nd ACM SIGSAC conference on computer
  and communications security}, pp.\  1322--1333, 2015.

\bibitem[He et~al.(2016)He, Zhang, Ren, and Sun]{he2016deep}
He, K., Zhang, X., Ren, S., and Sun, J.
\newblock Deep residual learning for image recognition.
\newblock In \emph{Proceedings of the IEEE conference on computer vision and
  pattern recognition}, pp.\  770--778, 2016.

\bibitem[Heo et~al.(2021)Heo, Yun, Han, Chun, Choe, and Oh]{heo2021rethinking}
Heo, B., Yun, S., Han, D., Chun, S., Choe, J., and Oh, S.~J.
\newblock Rethinking spatial dimensions of vision transformers.
\newblock \emph{arXiv preprint arXiv:2103.16302}, 2021.

\bibitem[Howard et~al.(2019)Howard, Sandler, Chu, Chen, Chen, Tan, Wang, Zhu,
  Pang, Vasudevan, et~al.]{howard2019searching}
Howard, A., Sandler, M., Chu, G., Chen, L.-C., Chen, B., Tan, M., Wang, W.,
  Zhu, Y., Pang, R., Vasudevan, V., et~al.
\newblock Searching for mobilenetv3.
\newblock In \emph{Proceedings of the IEEE/CVF International Conference on
  Computer Vision}, pp.\  1314--1324, 2019.

\bibitem[Huang et~al.(2017)Huang, Liu, Van Der~Maaten, and
  Weinberger]{huang2017densely}
Huang, G., Liu, Z., Van Der~Maaten, L., and Weinberger, K.~Q.
\newblock Densely connected convolutional networks.
\newblock In \emph{Proceedings of the IEEE conference on computer vision and
  pattern recognition}, pp.\  4700--4708, 2017.

\bibitem[Iandola et~al.(2016)Iandola, Han, Moskewicz, Ashraf, Dally, and
  Keutzer]{iandola2016squeezenet}
Iandola, F.~N., Han, S., Moskewicz, M.~W., Ashraf, K., Dally, W.~J., and
  Keutzer, K.
\newblock Squeezenet: Alexnet-level accuracy with 50x fewer parameters and< 0.5
  mb model size.
\newblock \emph{arXiv preprint arXiv:1602.07360}, 2016.

\bibitem[Ioffe \& Szegedy(2015)Ioffe and Szegedy]{ioffe2015batch}
Ioffe, S. and Szegedy, C.
\newblock Batch normalization: Accelerating deep network training by reducing
  internal covariate shift.
\newblock In \emph{International conference on machine learning}, pp.\
  448--456. PMLR, 2015.

\bibitem[Kingma \& Ba(2014)Kingma and Ba]{kingma2014adam}
Kingma, D.~P. and Ba, J.
\newblock Adam: A method for stochastic optimization.
\newblock \emph{arXiv preprint arXiv:1412.6980}, 2014.

\bibitem[Krizhevsky et~al.(2012)Krizhevsky, Sutskever, and
  Hinton]{krizhevsky2012imagenet}
Krizhevsky, A., Sutskever, I., and Hinton, G.~E.
\newblock Imagenet classification with deep convolutional neural networks.
\newblock \emph{Advances in neural information processing systems},
  25:\penalty0 1097--1105, 2012.

\bibitem[LeCun et~al.(1989)LeCun, Boser, Denker, Henderson, Howard, Hubbard,
  and Jackel]{lecun1989backpropagation}
LeCun, Y., Boser, B., Denker, J.~S., Henderson, D., Howard, R.~E., Hubbard, W.,
  and Jackel, L.~D.
\newblock Backpropagation applied to handwritten zip code recognition.
\newblock \emph{Neural computation}, 1\penalty0 (4):\penalty0 541--551, 1989.

\bibitem[Liu et~al.(2021{\natexlab{a}})Liu, Dai, So, and Le]{liu2021pay}
Liu, H., Dai, Z., So, D.~R., and Le, Q.~V.
\newblock Pay attention to mlps.
\newblock \emph{arXiv preprint arXiv:2105.08050}, 2021{\natexlab{a}}.

\bibitem[Liu et~al.(2021{\natexlab{b}})Liu, Lin, Cao, Hu, Wei, Zhang, Lin, and
  Guo]{liu2021swin}
Liu, Z., Lin, Y., Cao, Y., Hu, H., Wei, Y., Zhang, Z., Lin, S., and Guo, B.
\newblock Swin transformer: Hierarchical vision transformer using shifted
  windows.
\newblock \emph{arXiv preprint arXiv:2103.14030}, 2021{\natexlab{b}}.

\bibitem[Ma et~al.(2018)Ma, Zhang, Zheng, and Sun]{ma2018shufflenet}
Ma, N., Zhang, X., Zheng, H.-T., and Sun, J.
\newblock Shufflenet v2: Practical guidelines for efficient cnn architecture
  design.
\newblock In \emph{Proceedings of the European conference on computer vision
  (ECCV)}, pp.\  116--131, 2018.

\bibitem[Mahendran \& Vedaldi(2015)Mahendran and
  Vedaldi]{mahendran2015understanding}
Mahendran, A. and Vedaldi, A.
\newblock Understanding deep image representations by inverting them.
\newblock In \emph{Proceedings of the IEEE conference on computer vision and
  pattern recognition}, pp.\  5188--5196, 2015.

\bibitem[Mejia et~al.(2019)Mejia, Gamble, Hampel-Arias, Lomnitz, Lopatina,
  Tindall, and Barrios]{mejia2019robust}
Mejia, F.~A., Gamble, P., Hampel-Arias, Z., Lomnitz, M., Lopatina, N., Tindall,
  L., and Barrios, M.~A.
\newblock Robust or private? adversarial training makes models more vulnerable
  to privacy attacks.
\newblock \emph{arXiv preprint arXiv:1906.06449}, 2019.

\bibitem[Mordvintsev et~al.(2015)Mordvintsev, Olah, and
  Tyka]{mordvintsev2015inceptionism}
Mordvintsev, A., Olah, C., and Tyka, M.
\newblock Inceptionism: Going deeper into neural networks.
\newblock 2015.

\bibitem[Olah et~al.(2017)Olah, Mordvintsev, and Schubert]{olah2017feature}
Olah, C., Mordvintsev, A., and Schubert, L.
\newblock Feature visualization.
\newblock \emph{Distill}, 2017.
\newblock \doi{10.23915/distill.00007}.
\newblock https://distill.pub/2017/feature-visualization.

\bibitem[Opitz \& Maclin(1999)Opitz and Maclin]{opitz1999popular}
Opitz, D. and Maclin, R.
\newblock Popular ensemble methods: An empirical study.
\newblock \emph{Journal of artificial intelligence research}, 11:\penalty0
  169--198, 1999.

\bibitem[Paszke et~al.(2019)Paszke, Gross, Massa, Lerer, Bradbury, Chanan,
  Killeen, Lin, Gimelshein, Antiga, et~al.]{paszke2019pytorch}
Paszke, A., Gross, S., Massa, F., Lerer, A., Bradbury, J., Chanan, G., Killeen,
  T., Lin, Z., Gimelshein, N., Antiga, L., et~al.
\newblock Pytorch: An imperative style, high-performance deep learning library.
\newblock \emph{Advances in neural information processing systems},
  32:\penalty0 8026--8037, 2019.

\bibitem[Salimans et~al.(2016)Salimans, Goodfellow, Zaremba, Cheung, Radford,
  and Chen]{salimans2016improved}
Salimans, T., Goodfellow, I., Zaremba, W., Cheung, V., Radford, A., and Chen,
  X.
\newblock Improved techniques for training gans.
\newblock \emph{Advances in neural information processing systems},
  29:\penalty0 2234--2242, 2016.

\bibitem[Sandler et~al.(2018)Sandler, Howard, Zhu, Zhmoginov, and
  Chen]{sandler2018mobilenetv2}
Sandler, M., Howard, A., Zhu, M., Zhmoginov, A., and Chen, L.-C.
\newblock Mobilenetv2: Inverted residuals and linear bottlenecks.
\newblock In \emph{Proceedings of the IEEE conference on computer vision and
  pattern recognition}, pp.\  4510--4520, 2018.

\bibitem[Santurkar et~al.(2019)Santurkar, Ilyas, Tsipras, Engstrom, Tran, and
  Madry]{santurkar2019image}
Santurkar, S., Ilyas, A., Tsipras, D., Engstrom, L., Tran, B., and Madry, A.
\newblock Image synthesis with a single (robust) classifier.
\newblock \emph{Advances in Neural Information Processing Systems},
  32:\penalty0 1262--1273, 2019.

\bibitem[Shafahi et~al.(2019)Shafahi, Najibi, Ghiasi, Xu, Dickerson, Studer,
  Davis, Taylor, and Goldstein]{shafahi2019adversarial}
Shafahi, A., Najibi, M., Ghiasi, A., Xu, Z., Dickerson, J., Studer, C., Davis,
  L.~S., Taylor, G., and Goldstein, T.
\newblock Adversarial training for free!
\newblock \emph{arXiv preprint arXiv:1904.12843}, 2019.

\bibitem[Simonyan \& Zisserman(2014)Simonyan and Zisserman]{simonyan2014very}
Simonyan, K. and Zisserman, A.
\newblock Very deep convolutional networks for large-scale image recognition.
\newblock \emph{arXiv preprint arXiv:1409.1556}, 2014.

\bibitem[Simonyan et~al.(2014)Simonyan, Vedaldi, and
  Zisserman]{simonyan2014deep}
Simonyan, K., Vedaldi, A., and Zisserman, A.
\newblock Deep inside convolutional networks: Visualising image classification
  models and saliency maps, 2014.

\bibitem[Souri et~al.(2021)Souri, Goldblum, Fowl, Chellappa, and
  Goldstein]{souri2021sleeper}
Souri, H., Goldblum, M., Fowl, L., Chellappa, R., and Goldstein, T.
\newblock Sleeper agent: Scalable hidden trigger backdoors for neural networks
  trained from scratch.
\newblock \emph{arXiv preprint arXiv:2106.08970}, 2021.

\bibitem[Szegedy et~al.(2015)Szegedy, Liu, Jia, Sermanet, Reed, Anguelov,
  Erhan, Vanhoucke, and Rabinovich]{szegedy2015going}
Szegedy, C., Liu, W., Jia, Y., Sermanet, P., Reed, S., Anguelov, D., Erhan, D.,
  Vanhoucke, V., and Rabinovich, A.
\newblock Going deeper with convolutions.
\newblock In \emph{Proceedings of the IEEE conference on computer vision and
  pattern recognition}, pp.\  1--9, 2015.

\bibitem[Tan et~al.(2019)Tan, Chen, Pang, Vasudevan, Sandler, Howard, and
  Le]{tan2019mnasnet}
Tan, M., Chen, B., Pang, R., Vasudevan, V., Sandler, M., Howard, A., and Le,
  Q.~V.
\newblock Mnasnet: Platform-aware neural architecture search for mobile.
\newblock In \emph{Proceedings of the IEEE/CVF Conference on Computer Vision
  and Pattern Recognition}, pp.\  2820--2828, 2019.

\bibitem[Tolstikhin et~al.(2021)Tolstikhin, Houlsby, Kolesnikov, Beyer, Zhai,
  Unterthiner, Yung, Keysers, Uszkoreit, Lucic, et~al.]{tolstikhin2021mlp}
Tolstikhin, I., Houlsby, N., Kolesnikov, A., Beyer, L., Zhai, X., Unterthiner,
  T., Yung, J., Keysers, D., Uszkoreit, J., Lucic, M., et~al.
\newblock Mlp-mixer: An all-mlp architecture for vision.
\newblock \emph{arXiv preprint arXiv:2105.01601}, 2021.

\bibitem[Touvron et~al.(2021{\natexlab{a}})Touvron, Bojanowski, Caron, Cord,
  El-Nouby, Grave, Izacard, Joulin, Synnaeve, Verbeek, and
  Jégou]{touvron2021resmlp}
Touvron, H., Bojanowski, P., Caron, M., Cord, M., El-Nouby, A., Grave, E.,
  Izacard, G., Joulin, A., Synnaeve, G., Verbeek, J., and Jégou, H.
\newblock Resmlp: Feedforward networks for image classification with
  data-efficient training, 2021{\natexlab{a}}.

\bibitem[Touvron et~al.(2021{\natexlab{b}})Touvron, Cord, Douze, Massa,
  Sablayrolles, and Jegou]{pmlrv139touvron21a}
Touvron, H., Cord, M., Douze, M., Massa, F., Sablayrolles, A., and Jegou, H.
\newblock Training data-efficient image transformers \& distillation through
  attention.
\newblock In \emph{International Conference on Machine Learning}, volume 139,
  pp.\  10347--10357, July 2021{\natexlab{b}}.

\bibitem[Touvron et~al.(2021{\natexlab{c}})Touvron, Cord, Douze, Massa,
  Sablayrolles, and J{\'e}gou]{touvron2021training}
Touvron, H., Cord, M., Douze, M., Massa, F., Sablayrolles, A., and J{\'e}gou,
  H.
\newblock Training data-efficient image transformers \& distillation through
  attention.
\newblock In \emph{International Conference on Machine Learning}, pp.\
  10347--10357. PMLR, 2021{\natexlab{c}}.

\bibitem[Vaswani et~al.(2017)Vaswani, Shazeer, Parmar, Uszkoreit, Jones, Gomez,
  Kaiser, and Polosukhin]{vaswani2017attention}
Vaswani, A., Shazeer, N., Parmar, N., Uszkoreit, J., Jones, L., Gomez, A.~N.,
  Kaiser, {\L}., and Polosukhin, I.
\newblock Attention is all you need.
\newblock In \emph{Advances in neural information processing systems}, pp.\
  5998--6008, 2017.

\bibitem[Wightman(2019)]{rw2019timm}
Wightman, R.
\newblock Pytorch image models.
\newblock \url{https://github.com/rwightman/pytorch-image-models}, 2019.

\bibitem[Xie et~al.(2017)Xie, Girshick, Doll{\'a}r, Tu, and
  He]{xie2017aggregated}
Xie, S., Girshick, R., Doll{\'a}r, P., Tu, Z., and He, K.
\newblock Aggregated residual transformations for deep neural networks.
\newblock In \emph{Proceedings of the IEEE conference on computer vision and
  pattern recognition}, pp.\  1492--1500, 2017.

\bibitem[Xu et~al.(2021)Xu, Xu, Chang, and Tu]{xu2021co}
Xu, W., Xu, Y., Chang, T., and Tu, Z.
\newblock Co-scale conv-attentional image transformers.
\newblock \emph{arXiv preprint arXiv:2104.06399}, 2021.

\bibitem[Yin et~al.(2020)Yin, Molchanov, Alvarez, Li, Mallya, Hoiem, Jha, and
  Kautz]{yin2020dreaming}
Yin, H., Molchanov, P., Alvarez, J.~M., Li, Z., Mallya, A., Hoiem, D., Jha,
  N.~K., and Kautz, J.
\newblock Dreaming to distill: Data-free knowledge transfer via deepinversion.
\newblock In \emph{Proceedings of the IEEE/CVF Conference on Computer Vision
  and Pattern Recognition}, pp.\  8715--8724, 2020.

\bibitem[Zagoruyko \& Komodakis(2016)Zagoruyko and
  Komodakis]{zagoruyko2016wide}
Zagoruyko, S. and Komodakis, N.
\newblock Wide residual networks.
\newblock \emph{arXiv preprint arXiv:1605.07146}, 2016.

\bibitem[Zeiler \& Fergus(2014)Zeiler and Fergus]{zeiler2014visualizing}
Zeiler, M.~D. and Fergus, R.
\newblock Visualizing and understanding convolutional networks.
\newblock In \emph{European conference on computer vision}, pp.\  818--833.
  Springer, 2014.

\end{thebibliography}
\bibliographystyle{icml2022}

\clearpage

\appendix
\onecolumn
\section{Additional Results}

\subsection{Ablation Study for $\alpha$ and $\beta$}

Figure \ref{fig:app:mean_var} shows the results of \methodname{} when varying the values of $\alpha$ and $\beta$, which determine the intervals from which the ColorShift constants are randomly drawn. Based on this and similar experiments, we permanently fix these parameters to $\alpha = \beta = 1.0$ for all other \methodname{} experiments, and find that these values indeed transfer well to other models.

\def \varmeanvar{1\linewidth} 
\begin{figure}[!h]
    \centering
    \setlength\tabcolsep{1.5pt}
    \begin{tabularx}{\linewidth}{cYYYYYYY}
        & $\beta=0$ & $\beta=0.1$ & $\beta=0.5$ & $\beta=1.0$ & $\beta=2.0$ & $\beta=4.0$ & $\beta=8.0$ \\

         \raisebox{0.4\totalheight}{\rotatebox[origin=lB]{90}{$\alpha=0$} } & 
        \includegraphics[width=\varmeanvar]{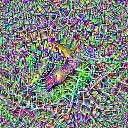}  &
        \includegraphics[width=\varmeanvar]{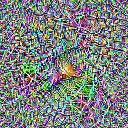}  &
        \includegraphics[width=\varmeanvar]{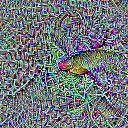}  &
        \includegraphics[width=\varmeanvar]{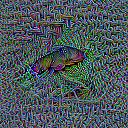}  &
        \includegraphics[width=\varmeanvar]{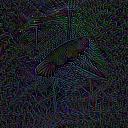}  &
        \includegraphics[width=\varmeanvar]{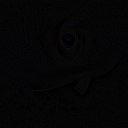}  &
        \includegraphics[width=\varmeanvar]{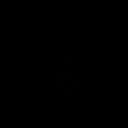}  \\
        
        \raisebox{0.4\totalheight}{\rotatebox[origin=lB]{90}{$\alpha=0.1$} } & 
        \includegraphics[width=\varmeanvar]{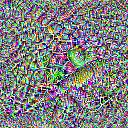}  &
        \includegraphics[width=\varmeanvar]{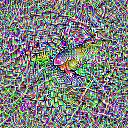}  &
        \includegraphics[width=\varmeanvar]{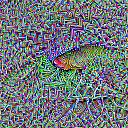}  &
        \includegraphics[width=\varmeanvar]{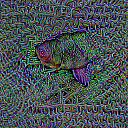}  &
        \includegraphics[width=\varmeanvar]{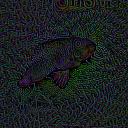}  &
        \includegraphics[width=\varmeanvar]{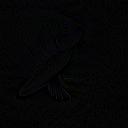}  &
        \includegraphics[width=\varmeanvar]{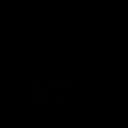}  \\
        
        \raisebox{0.4\totalheight}{\rotatebox[origin=lB]{90}{$\alpha=0.5$} } & 
        \includegraphics[width=\varmeanvar]{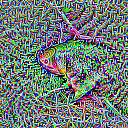}  &
        \includegraphics[width=\varmeanvar]{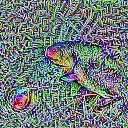}  &
        \includegraphics[width=\varmeanvar]{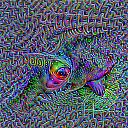}  &
        \includegraphics[width=\varmeanvar]{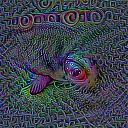}  &
        \includegraphics[width=\varmeanvar]{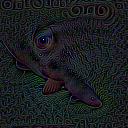}  &
        \includegraphics[width=\varmeanvar]{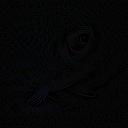}  &
        \includegraphics[width=\varmeanvar]{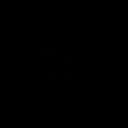}  \\
        
        \raisebox{0.4\totalheight}{\rotatebox[origin=lB]{90}{$\alpha=1.0$} } & 
        \includegraphics[width=\varmeanvar]{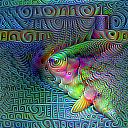}  &
        \includegraphics[width=\varmeanvar]{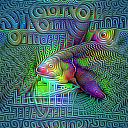}  &
        \includegraphics[width=\varmeanvar]{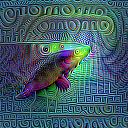}  &
        \includegraphics[width=\varmeanvar]{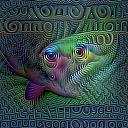}  &
        \includegraphics[width=\varmeanvar]{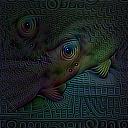}  &
        \includegraphics[width=\varmeanvar]{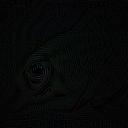}  &
        \includegraphics[width=\varmeanvar]{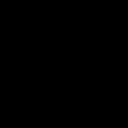}  \\
        
        \raisebox{0.4\totalheight}{\rotatebox[origin=lB]{90}{$\alpha=2.0$} } & 
        \includegraphics[width=\varmeanvar]{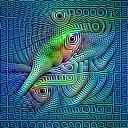}  &
        \includegraphics[width=\varmeanvar]{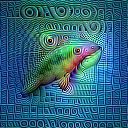}  &
        \includegraphics[width=\varmeanvar]{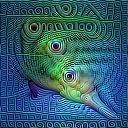}  &
        \includegraphics[width=\varmeanvar]{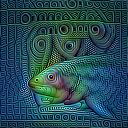}  &
        \includegraphics[width=\varmeanvar]{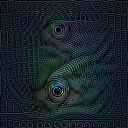}  &
        \includegraphics[width=\varmeanvar]{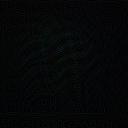}  &
        \includegraphics[width=\varmeanvar]{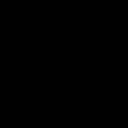}  \\
        
        \raisebox{0.4\totalheight}{\rotatebox[origin=lB]{90}{$\alpha=4.0$} } & 
        \includegraphics[width=\varmeanvar]{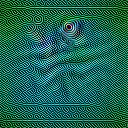}  &
        \includegraphics[width=\varmeanvar]{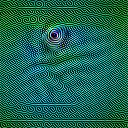}  &
        \includegraphics[width=\varmeanvar]{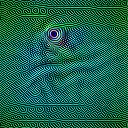}  &
        \includegraphics[width=\varmeanvar]{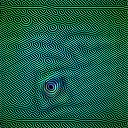}  &
        \includegraphics[width=\varmeanvar]{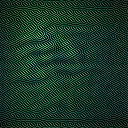}  &
        \includegraphics[width=\varmeanvar]{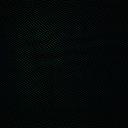}  &
        \includegraphics[width=\varmeanvar]{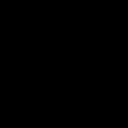}  \\
        
        \raisebox{0.4\totalheight}{\rotatebox[origin=lB]{90}{$\alpha=8.0$} } & 
        \includegraphics[width=\varmeanvar]{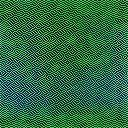}  &
        \includegraphics[width=\varmeanvar]{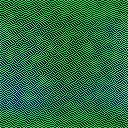}  &
        \includegraphics[width=\varmeanvar]{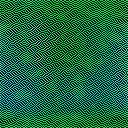}  &
        \includegraphics[width=\varmeanvar]{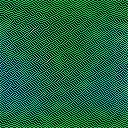}  &
        \includegraphics[width=\varmeanvar]{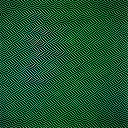}  &
        \includegraphics[width=\varmeanvar]{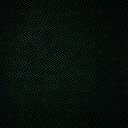}  &
        \includegraphics[width=\varmeanvar]{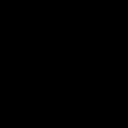}  \\

    \end{tabularx}
    \caption{ Effect of $\alpha$ and $\beta$ on the quality of the images generated by \text{\methodname} from a naturally-trained ResNet-50 for the Tench class.}
    \label{fig:app:mean_var}
\end{figure}

\subsection{Insensitivity to TV regularization}

Figure \ref{fig:app:tv} shows additional results on the effect of ColorShift on the sensitivity to the weight of TV regularization when inverting a robust model, complementing Figure \ref{fig:tv}. As in the earlier figure, we observe that certain values of $\lambda_{TV}$ may produce noisy or blurred images when not using ColorShift, whereas the ColorShift results are quite stable.

\def \vartv{1\linewidth} 
\begin{figure}[!h]
    \centering
    \setlength\tabcolsep{1.5pt}
    \begin{tabularx}{0.9\linewidth}{cYYYYYY}
            
        $log(\lambda_{tv}):$ & $-9$ & $-8$ & $-7$ & $-6$ & $-5$ & $-4$ \\ 
        \raisebox{3\totalheight}{w/ \text{\augnameshort}} &
        \includegraphics[width=\vartv]{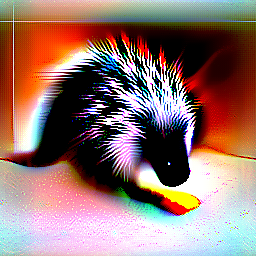}  &
        \includegraphics[width=\vartv]{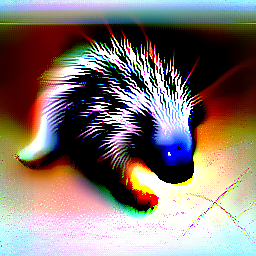}  &
        \includegraphics[width=\vartv]{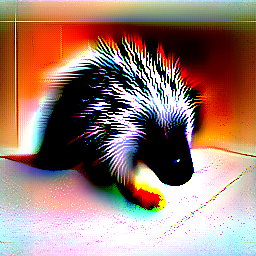}  &
        \includegraphics[width=\vartv]{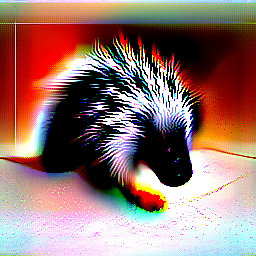}  &
        \includegraphics[width=\vartv]{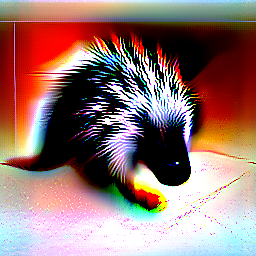}  &
        \includegraphics[width=\vartv]{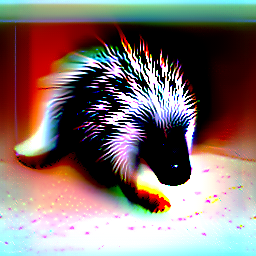}  \\        \raisebox{3\totalheight}{w/o \text{\augnameshort}} &
        \includegraphics[width=\vartv]{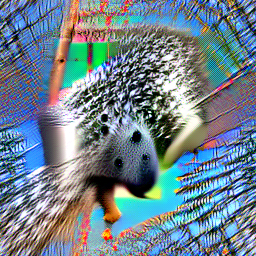}  &
        \includegraphics[width=\vartv]{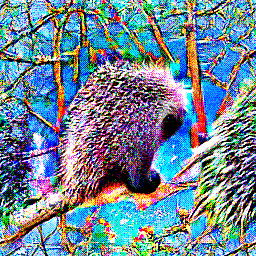}  &
        \includegraphics[width=\vartv]{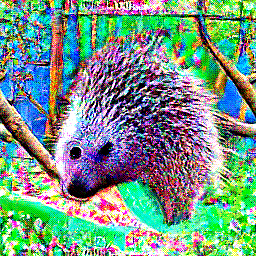}  &
        \includegraphics[width=\vartv]{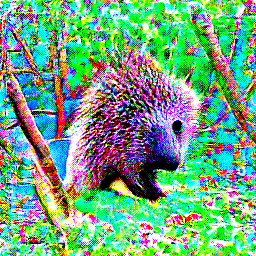}  &
        \includegraphics[width=\vartv]{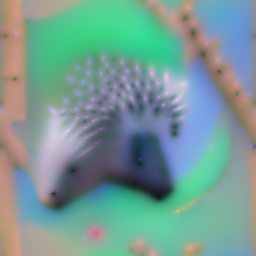}  &
        \includegraphics[width=\vartv]{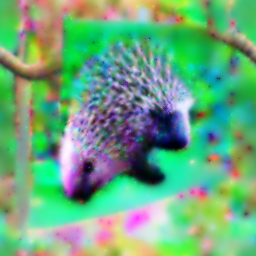}  \\  \hline

        \raisebox{3\totalheight}{w/ \text{\augnameshort}} &
        \includegraphics[width=\vartv]{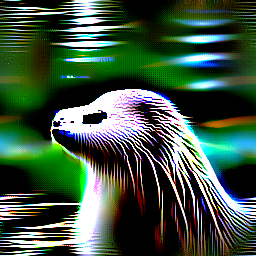}  &
        \includegraphics[width=\vartv]{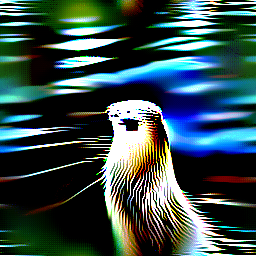}  &
        \includegraphics[width=\vartv]{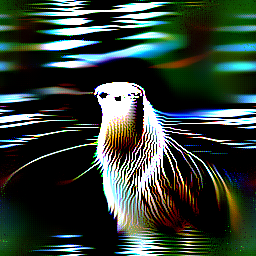}  &
        \includegraphics[width=\vartv]{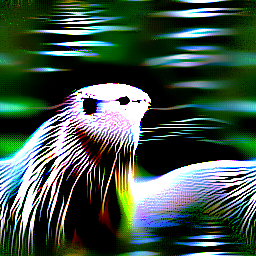}  &
        \includegraphics[width=\vartv]{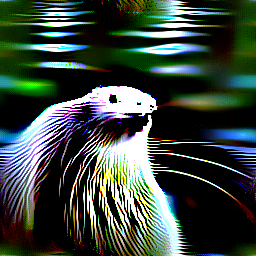}  &
        \includegraphics[width=\vartv]{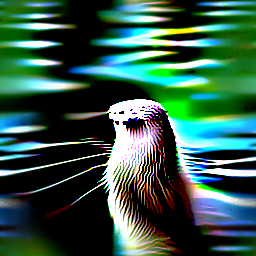}  \\        \raisebox{3\totalheight}{w/o \text{\augnameshort}} &
        \includegraphics[width=\vartv]{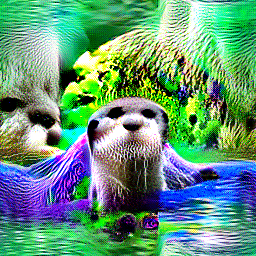}  &
        \includegraphics[width=\vartv]{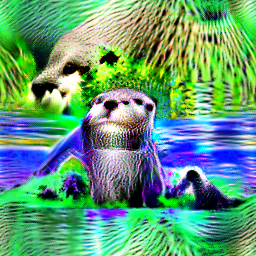}  &
        \includegraphics[width=\vartv]{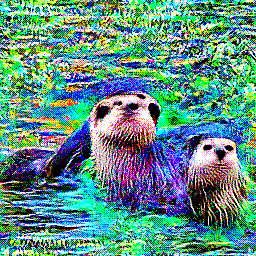}  &
        \includegraphics[width=\vartv]{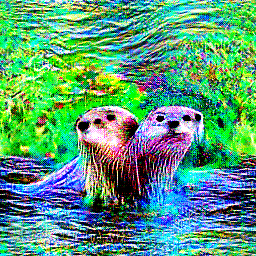}  &
        \includegraphics[width=\vartv]{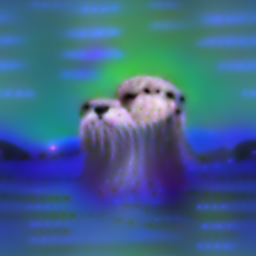}  &
        \includegraphics[width=\vartv]{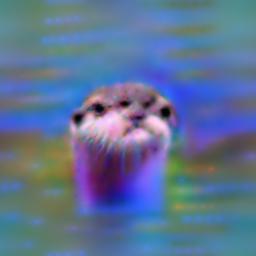}  \\  \hline

        \raisebox{3\totalheight}{w/ \text{\augnameshort}} &
        \includegraphics[width=\vartv]{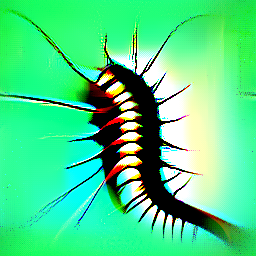}  &
        \includegraphics[width=\vartv]{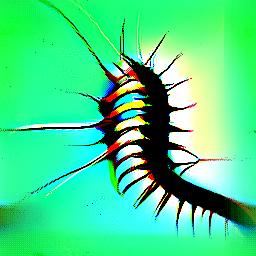}  &
        \includegraphics[width=\vartv]{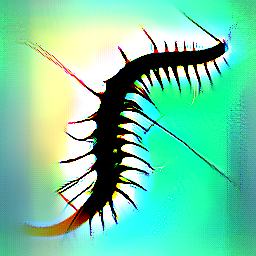}  &
        \includegraphics[width=\vartv]{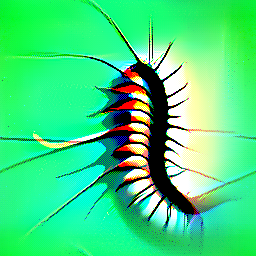}  &
        \includegraphics[width=\vartv]{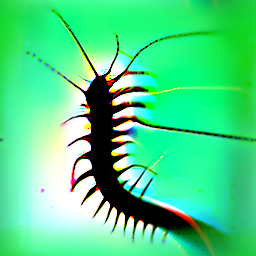}  &
        \includegraphics[width=\vartv]{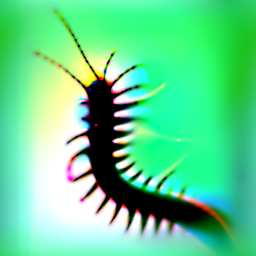}  \\        \raisebox{3\totalheight}{w/o \text{\augnameshort}} &
        \includegraphics[width=\vartv]{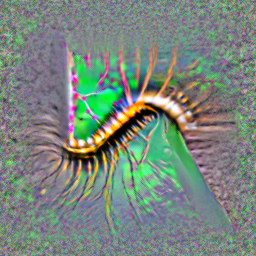}  &
        \includegraphics[width=\vartv]{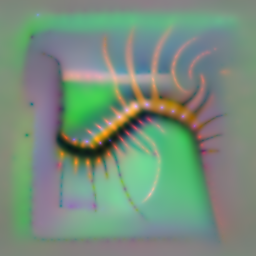}  &
        \includegraphics[width=\vartv]{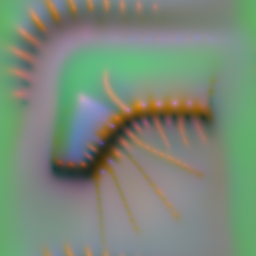}  &
        \includegraphics[width=\vartv]{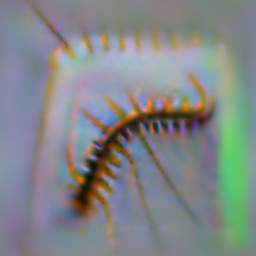}  &
        \includegraphics[width=\vartv]{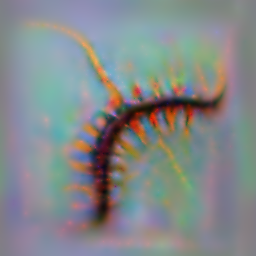}  &
        \includegraphics[width=\vartv]{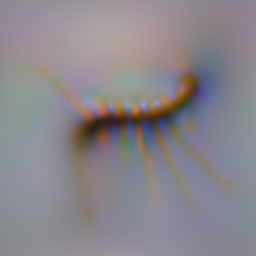}  \\  \hline
        
        \raisebox{3\totalheight}{w/ \text{\augnameshort}} &
        \includegraphics[width=\vartv]{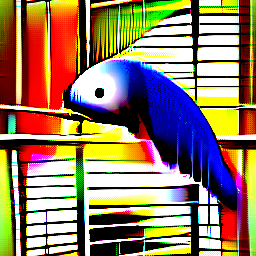}  &
        \includegraphics[width=\vartv]{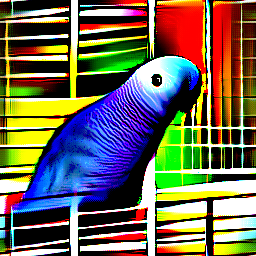}  &
        \includegraphics[width=\vartv]{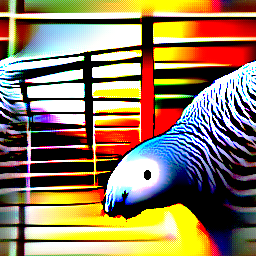}  &
        \includegraphics[width=\vartv]{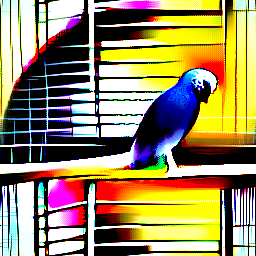}  &
        \includegraphics[width=\vartv]{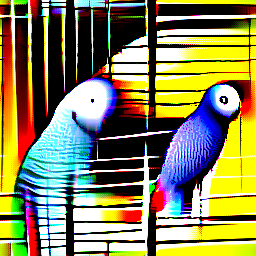}  &
        \includegraphics[width=\vartv]{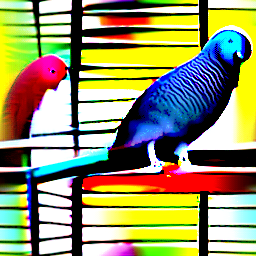}  \\        \raisebox{3\totalheight}{w/o \text{\augnameshort}} &
        \includegraphics[width=\vartv]{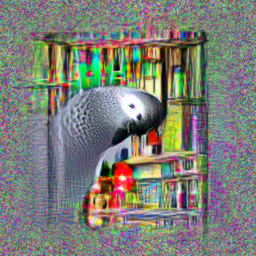}  &
        \includegraphics[width=\vartv]{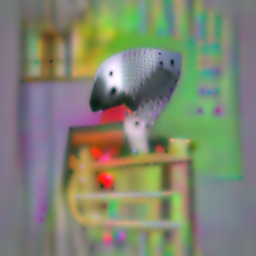}  &
        \includegraphics[width=\vartv]{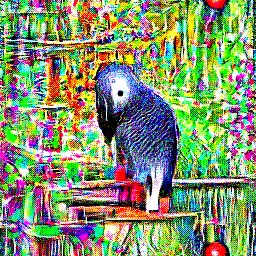}  &
        \includegraphics[width=\vartv]{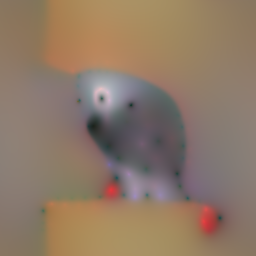}  &
        \includegraphics[width=\vartv]{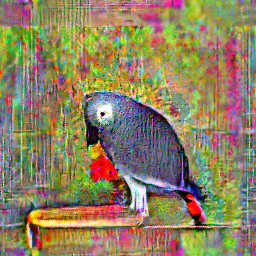}  &
        \includegraphics[width=\vartv]{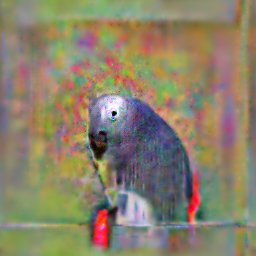}  \\ 

    \end{tabularx}
    \caption{ Effect of TV with and without \text{\augname}. With \text{\augname} it is clear that there is no need for hyper-parameter tuning for parameters such as TV. Images from the robust ResNet-50.}
    \label{fig:app:tv}
\end{figure}

\clearpage
\subsection{Effect of Centering}
Figures \ref{fig:app:robust_centering} and \ref{fig:app:nat_centering} show the effect of \emph{centering} on inverting a robust and natural model, respectively.

\def \varcenterapp{\linewidth} 
\def \varcenterappx{0pt}
\begin{figure}[h]
    \centering
    \setlength\tabcolsep{1.5pt}
    \begin{tabularx}{0.9\linewidth}{YYYYYYYY}
        
        Cen & Not Cen & Cen & Not Cen & Cen & Not Cen & Cen & Not Cen \\
            
        \raisebox{\varcenterappx}{\includegraphics[width=\varcenterapp]{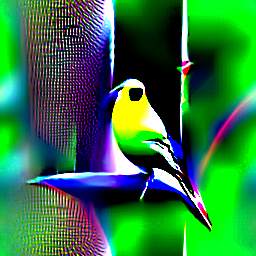} } &
        \raisebox{\varcenterappx}{\includegraphics[width=\varcenterapp]{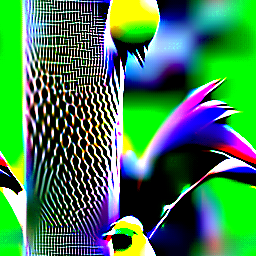} } &
        \raisebox{\varcenterappx}{\includegraphics[width=\varcenterapp]{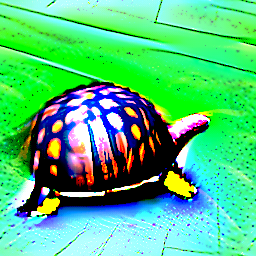} } &
        \raisebox{\varcenterappx}{\includegraphics[width=\varcenterapp]{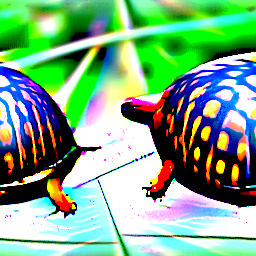} } &
        \raisebox{\varcenterappx}{\includegraphics[width=\varcenterapp]{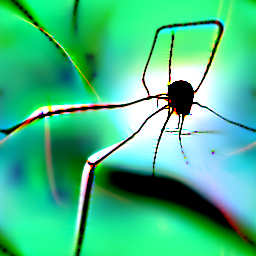} } &
        \raisebox{\varcenterappx}{\includegraphics[width=\varcenterapp]{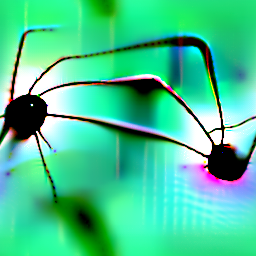} } &
        \raisebox{\varcenterappx}{\includegraphics[width=\varcenterapp]{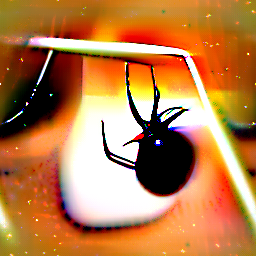} } &
        \raisebox{\varcenterappx}{\includegraphics[width=\varcenterapp]{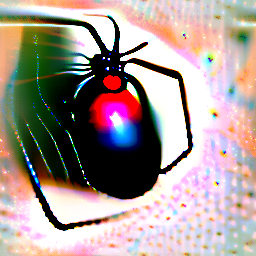} } \\ 
        
        \multicolumn{2}{c}{Gold Finch} & \multicolumn{2}{c}{Box Turtle} &  \multicolumn{2}{c}{Harvestman} & \multicolumn{2}{c}{Black Widow} \\

        \raisebox{\varcenterappx}{\includegraphics[width=\varcenterapp]{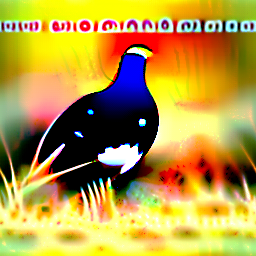} } &
        \raisebox{\varcenterappx}{\includegraphics[width=\varcenterapp]{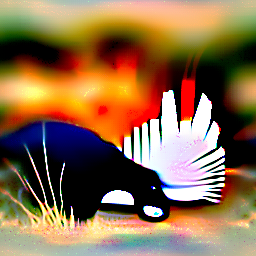} } &
        \raisebox{\varcenterappx}{\includegraphics[width=\varcenterapp]{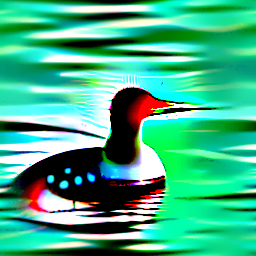} } &
        \raisebox{\varcenterappx}{\includegraphics[width=\varcenterapp]{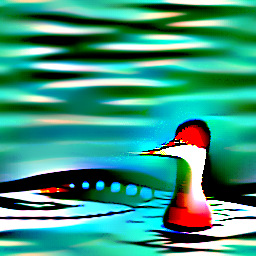} } &
        \raisebox{\varcenterappx}{\includegraphics[width=\varcenterapp]{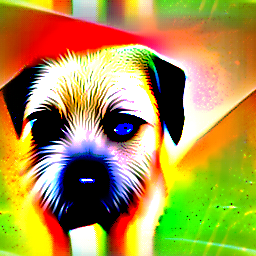} } &
        \raisebox{\varcenterappx}{\includegraphics[width=\varcenterapp]{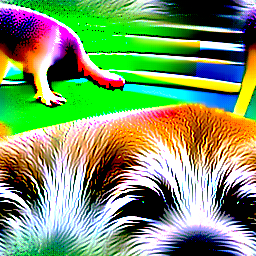} } &
        \raisebox{\varcenterappx}{\includegraphics[width=\varcenterapp]{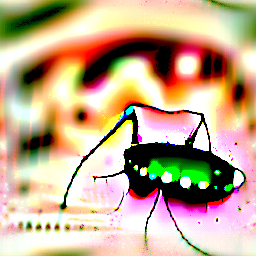} } &
        \raisebox{\varcenterappx}{\includegraphics[width=\varcenterapp]{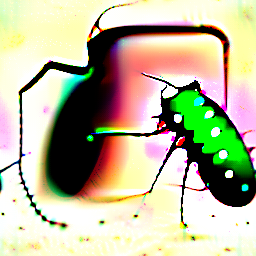} } \\ 
        
        \multicolumn{2}{c}{Black Grouse} & \multicolumn{2}{c}{Mergus Serrator} &  \multicolumn{2}{c}{Border Terrier} & \multicolumn{2}{c}{Tiger Beetle} \\ 

        \raisebox{\varcenterappx}{\includegraphics[width=\varcenterapp]{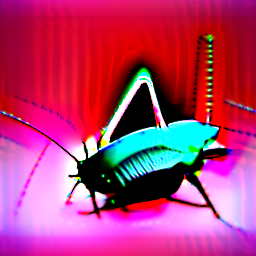} } &
        \raisebox{\varcenterappx}{\includegraphics[width=\varcenterapp]{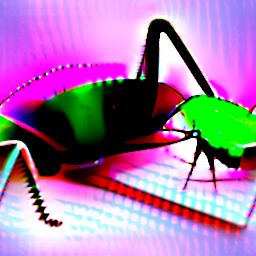} } &
        \raisebox{\varcenterappx}{\includegraphics[width=\varcenterapp]{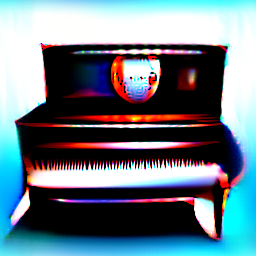} } &
        \raisebox{\varcenterappx}{\includegraphics[width=\varcenterapp]{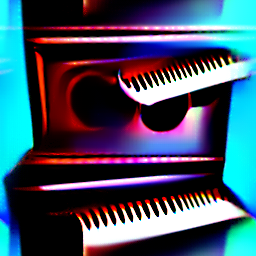} } &
        \raisebox{\varcenterappx}{\includegraphics[width=\varcenterapp]{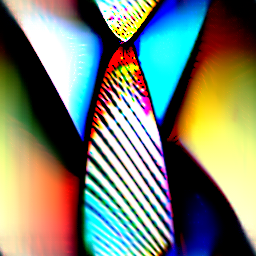} } &
        \raisebox{\varcenterappx}{\includegraphics[width=\varcenterapp]{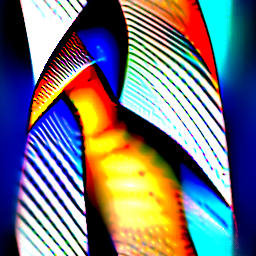} } &
        \raisebox{\varcenterappx}{\includegraphics[width=\varcenterapp]{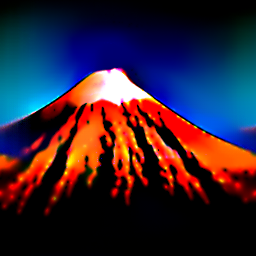} } &
        \raisebox{\varcenterappx}{\includegraphics[width=\varcenterapp]{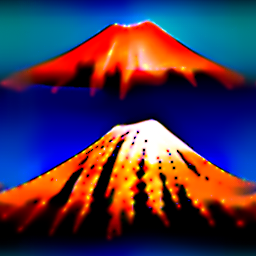} }  \\
        
        \multicolumn{2}{c}{Cricket} & \multicolumn{2}{c}{Upright Piano} &  \multicolumn{2}{c}{Windsor Tie} & \multicolumn{2}{c}{Volcano} \\ 
            
    \end{tabularx}
    \caption{ Effect of using centering vs not using centering for a robust ResNet-50.  }
    \label{fig:app:robust_centering}
\end{figure}

\begin{figure}[!h]
    \centering
    \setlength\tabcolsep{1.5pt}
    \begin{tabularx}{0.9\linewidth}{YYYYYYYY}
        Cen & Not Cen & Cen & Not Cen & Cen & Not Cen & Cen & Not Cen \\
            
        \raisebox{\varcenterappx}{\includegraphics[width=\varcenterapp]{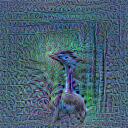}  } &
        \raisebox{\varcenterappx}{\includegraphics[width=\varcenterapp]{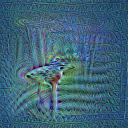}  } &
        \raisebox{\varcenterappx}{\includegraphics[width=\varcenterapp]{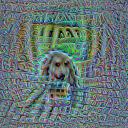}  } &
        \raisebox{\varcenterappx}{\includegraphics[width=\varcenterapp]{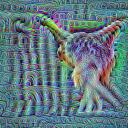}  } &
        \raisebox{\varcenterappx}{\includegraphics[width=\varcenterapp]{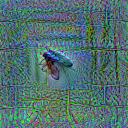}  } &
        \raisebox{\varcenterappx}{\includegraphics[width=\varcenterapp]{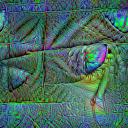}  } &
        \raisebox{\varcenterappx}{\includegraphics[width=\varcenterapp]{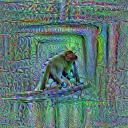}  } &
        \raisebox{\varcenterappx}{\includegraphics[width=\varcenterapp]{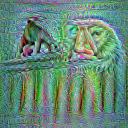} }  \\

        \multicolumn{2}{c}{Bustard} & \multicolumn{2}{c}{Otterhound} &  \multicolumn{2}{c}{Fly} & \multicolumn{2}{c}{Macaque} \\

        \raisebox{\varcenterappx}{\includegraphics[width=\varcenterapp]{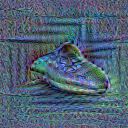}  } &
        \raisebox{\varcenterappx}{\includegraphics[width=\varcenterapp]{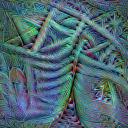}  } &
        \raisebox{\varcenterappx}{\includegraphics[width=\varcenterapp]{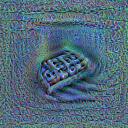}  } &
        \raisebox{\varcenterappx}{\includegraphics[width=\varcenterapp]{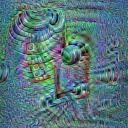}  } &
        \raisebox{\varcenterappx}{\includegraphics[width=\varcenterapp]{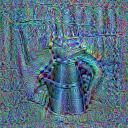}  } &
        \raisebox{\varcenterappx}{\includegraphics[width=\varcenterapp]{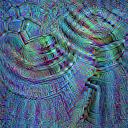}  } &
        \raisebox{\varcenterappx}{\includegraphics[width=\varcenterapp]{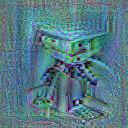}  } &
        \raisebox{\varcenterappx}{\includegraphics[width=\varcenterapp]{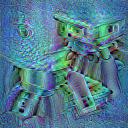} }  \\

        \multicolumn{2}{c}{Clog} & \multicolumn{2}{c}{Combination Lock} &  \multicolumn{2}{c}{Coffeepot} & \multicolumn{2}{c}{Espresso Maker} \\ 

        \raisebox{\varcenterappx}{\includegraphics[width=\varcenterapp]{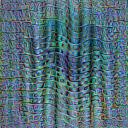}  } &
        \raisebox{\varcenterappx}{\includegraphics[width=\varcenterapp]{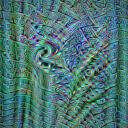}  } &
        \raisebox{\varcenterappx}{\includegraphics[width=\varcenterapp]{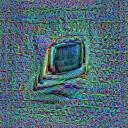}  } &
        \raisebox{\varcenterappx}{\includegraphics[width=\varcenterapp]{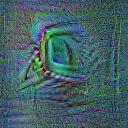}  } &
        \raisebox{\varcenterappx}{\includegraphics[width=\varcenterapp]{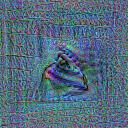}  } &
        \raisebox{\varcenterappx}{\includegraphics[width=\varcenterapp]{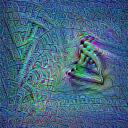}  } &
        \raisebox{\varcenterappx}{\includegraphics[width=\varcenterapp]{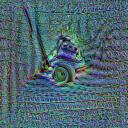}  } &
        \raisebox{\varcenterappx}{\includegraphics[width=\varcenterapp]{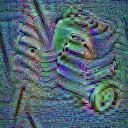} }  \\

        \multicolumn{2}{c}{Shower Curtain} & \multicolumn{2}{c}{TV} &  \multicolumn{2}{c}{Iron} & \multicolumn{2}{c}{Mower} \\ 

    \end{tabularx}
    \caption{ Effect of using centering vs not using centering for a naturally-trained ResNet-50. }
    \label{fig:app:nat_centering}
\end{figure}

\subsection{Effect of Ensemble Size}
\begin{figure}[!h]
    \centering
    \setlength\tabcolsep{1.5pt}
    \begin{tabularx}{0.95\linewidth}{cYYYYYYY}
            
         & $e=1$ & $e=2$ & $e=4$ & $e=8$ & $e=16$ & $e=32$ & $e=64$ \\ 

        \raisebox{1.2\totalheight}{\rotatebox[origin=lB]{90}{403} } &
        \includegraphics[width=\linewidth]{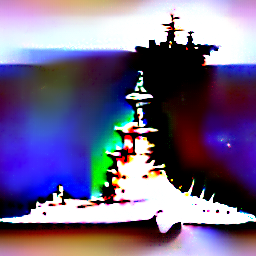} &
        \includegraphics[width=\linewidth]{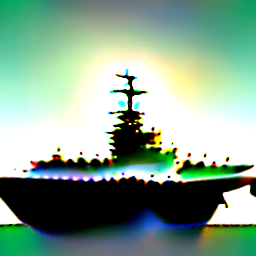} &
        \includegraphics[width=\linewidth]{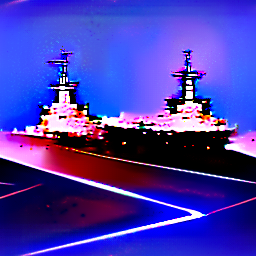} &
        \includegraphics[width=\linewidth]{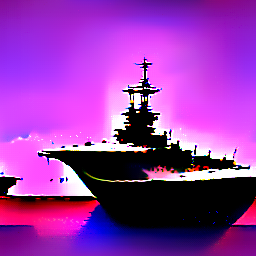} &
        \includegraphics[width=\linewidth]{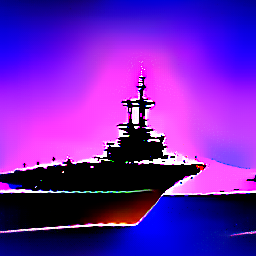} &
        \includegraphics[width=\linewidth]{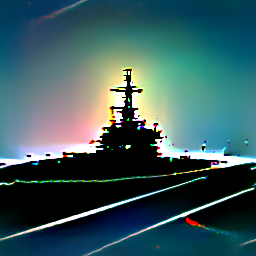} &
        \includegraphics[width=\linewidth]{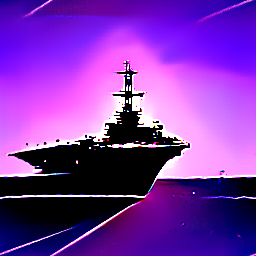} \\
        
        \raisebox{1.2\totalheight}{\rotatebox[origin=lB]{90}{283} } &
        \includegraphics[width=\linewidth]{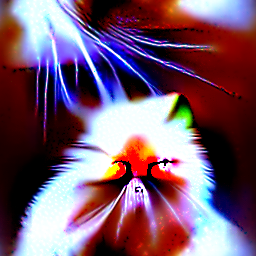} &
        \includegraphics[width=\linewidth]{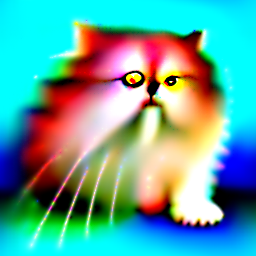} &
        \includegraphics[width=\linewidth]{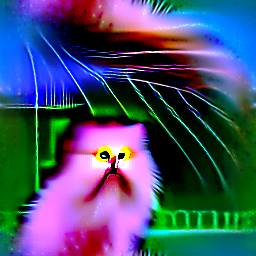} &
        \includegraphics[width=\linewidth]{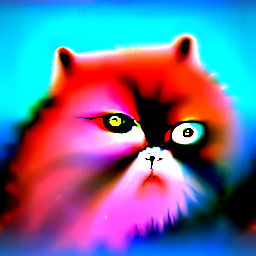} &
        \includegraphics[width=\linewidth]{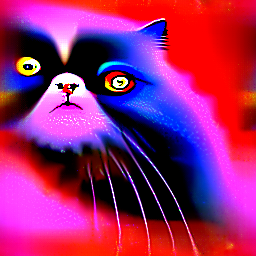} &
        \includegraphics[width=\linewidth]{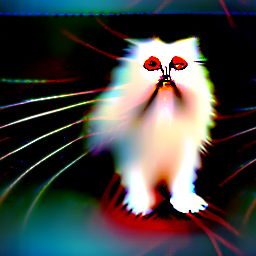} &
        \includegraphics[width=\linewidth]{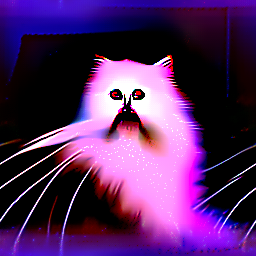} \\
        \raisebox{1.2\totalheight}{\rotatebox[origin=lB]{90}{449} } &
        \includegraphics[width=\linewidth]{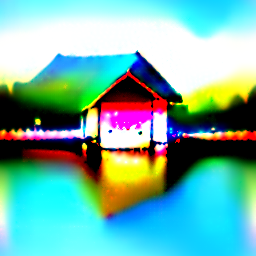} &
        \includegraphics[width=\linewidth]{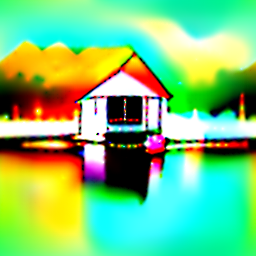} &
        \includegraphics[width=\linewidth]{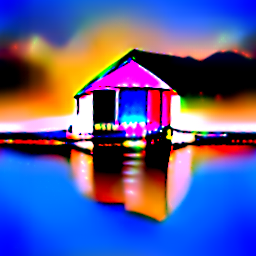} &
        \includegraphics[width=\linewidth]{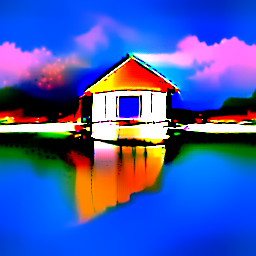} &
        \includegraphics[width=\linewidth]{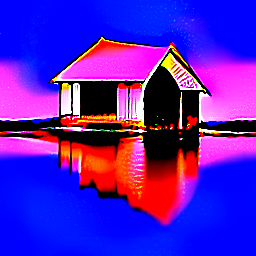} &
        \includegraphics[width=\linewidth]{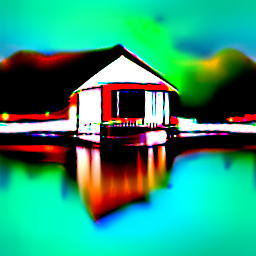} &
        \includegraphics[width=\linewidth]{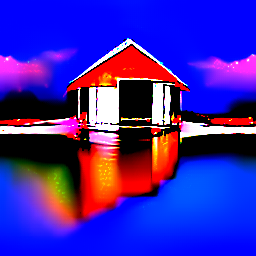} \\
        
        \raisebox{1.2\totalheight}{\rotatebox[origin=lB]{90}{460} } &
        \includegraphics[width=\linewidth]{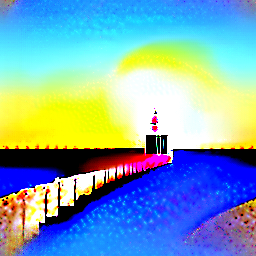} &
        \includegraphics[width=\linewidth]{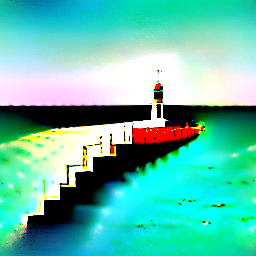} &
        \includegraphics[width=\linewidth]{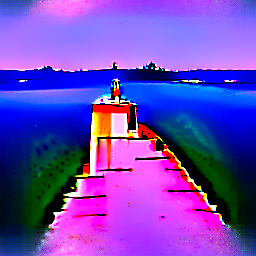} &
        \includegraphics[width=\linewidth]{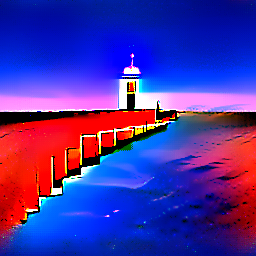} &
        \includegraphics[width=\linewidth]{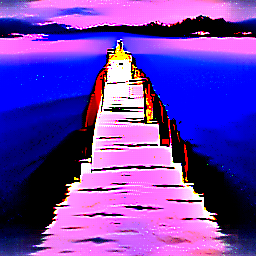} &
        \includegraphics[width=\linewidth]{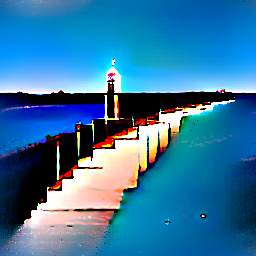} &
        \includegraphics[width=\linewidth]{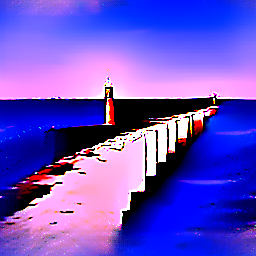} \\
        
        \raisebox{1.2\totalheight}{\rotatebox[origin=lB]{90}{558} } &
        \includegraphics[width=\linewidth]{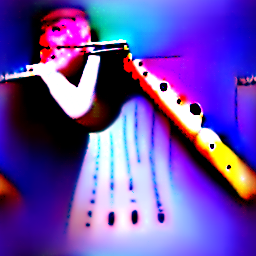} &
        \includegraphics[width=\linewidth]{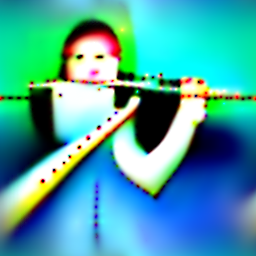} &
        \includegraphics[width=\linewidth]{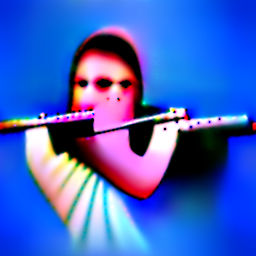} &
        \includegraphics[width=\linewidth]{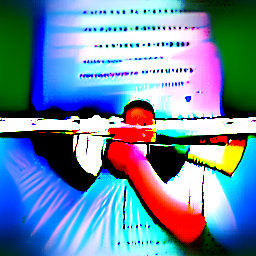} &
        \includegraphics[width=\linewidth]{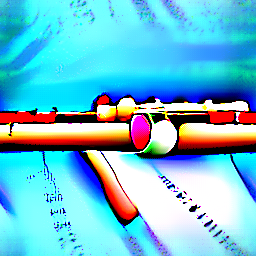} &
        \includegraphics[width=\linewidth]{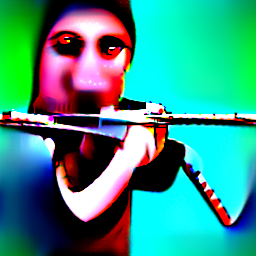} &
        \includegraphics[width=\linewidth]{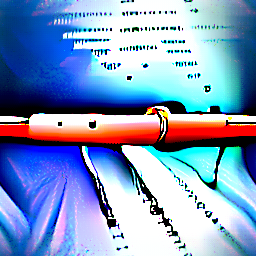} \\
        
        \raisebox{1.2\totalheight}{\rotatebox[origin=lB]{90}{802} } &
        \includegraphics[width=\linewidth]{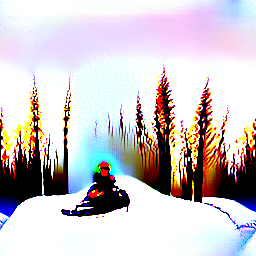} &
        \includegraphics[width=\linewidth]{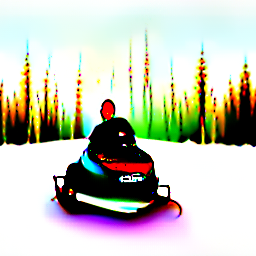} &
        \includegraphics[width=\linewidth]{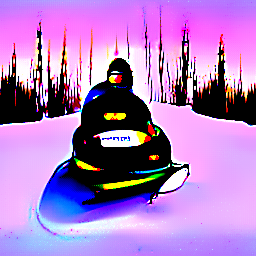} &
        \includegraphics[width=\linewidth]{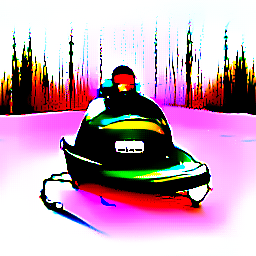} &
        \includegraphics[width=\linewidth]{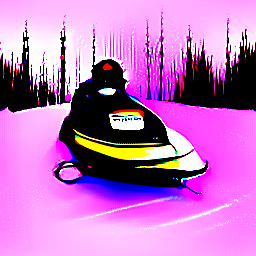} &
        \includegraphics[width=\linewidth]{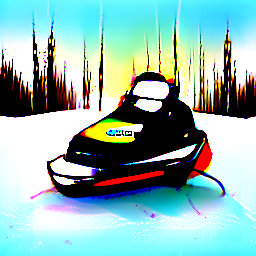} &
        \includegraphics[width=\linewidth]{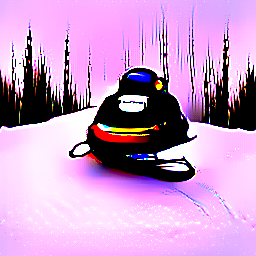} \\
        
        \raisebox{1.2\totalheight}{\rotatebox[origin=lB]{90}{834} } &
        \includegraphics[width=\linewidth]{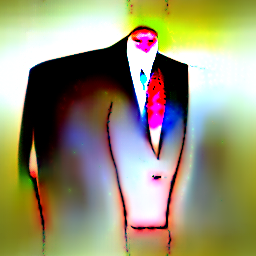} &
        \includegraphics[width=\linewidth]{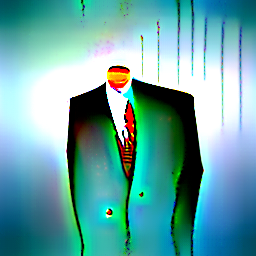} &
        \includegraphics[width=\linewidth]{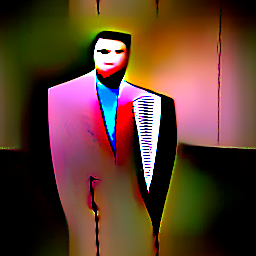} &
        \includegraphics[width=\linewidth]{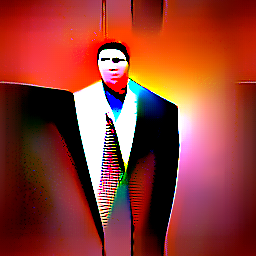} &
        \includegraphics[width=\linewidth]{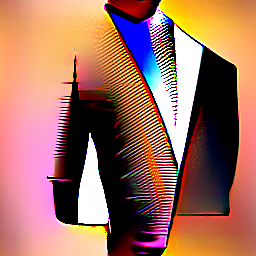} &
        \includegraphics[width=\linewidth]{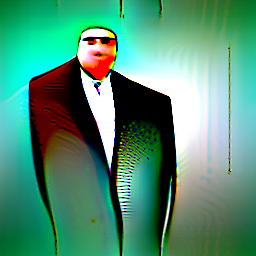} &
        \includegraphics[width=\linewidth]{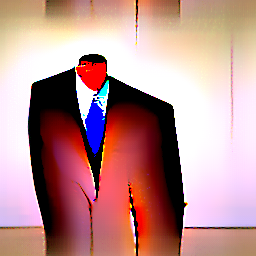} \\

    \end{tabularx}
    \caption{ Effect of ensemble size when inverting a robust ResNet-50. Even small values of $e$ give reasonably good results, but increasing $e$ tends to give slight improvement.}
        \label{fig:app:batch_size}
\end{figure}

Figure \ref{fig:app:batch_size} gives additional results to those in figure \ref{fig:batch_size} for the effect of ensemble size on inversion.
\clearpage

\subsection{Effect of using other Augmentations} \label{app:other_augs}
We used 4 random augmentations other than ColorShift to make comparisons. We used augmentations used in \cite{chen2020improved} with modifications. We use PyTorch \citep{paszke2019pytorch} notation to describe this part. We used \emph{RandomHorizontalFlip} with 0.5 probability. We used \emph{RandomResizedCrop} with scale [0.7, 1.], and ratio [0.75, and 1.33]. With applied \emph{ColorJitter} with 0.8 probability, and brightness, contrast, saturation, and hue of (0.4, 0.4, 0.4, 0.1), respectively. We used \emph{RandomGrayscale} with 0.2 probability. For this experiment, we do apply data normalization before feeding the input to the network. This is different than the regular experiment setting that we use for the robust model (see appendix \ref{app:robust-setting}). The reason is that not having data normalization is similar to using ColorShift (it changes the data distribution which the model expects as an input).

\def \varcenterapp{0.1195\linewidth} 
\def \varcenterappx{0pt}
\begin{figure}[!h]
    \centering
    \setlength\tabcolsep{1.5pt}
    \begin{tabularx}{\linewidth}{cYYYYYYY}
         & No Aug & Flip & Crop & Gray & Color Jitter & ColorShift \\ 
             
        \raisebox{0.1\totalheight}{\rotatebox[origin=lB]{90}{target=140} } & 
        \includegraphics[width=\linewidth]{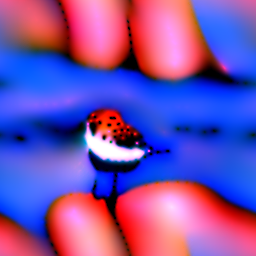} &
        \includegraphics[width=\linewidth]{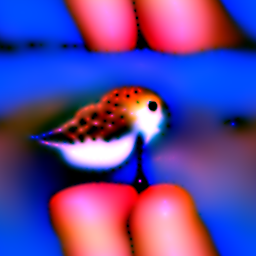} &
        \includegraphics[width=\linewidth]{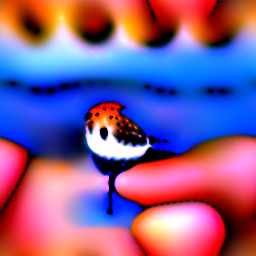} &
        \includegraphics[width=\linewidth]{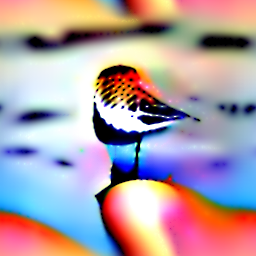} &
        \includegraphics[width=\linewidth]{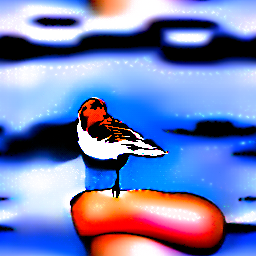} &
        \includegraphics[width=\linewidth]{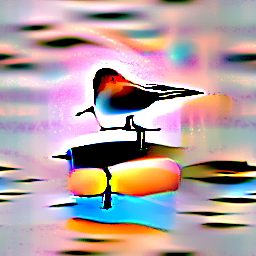} \\
        
        \raisebox{0.25\totalheight}{\rotatebox[origin=lB]{90}{target=295} } & 
        \includegraphics[width=\linewidth]{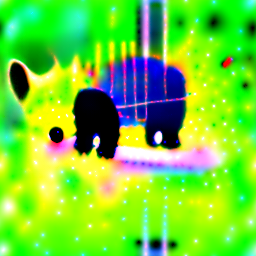} &
        \includegraphics[width=\linewidth]{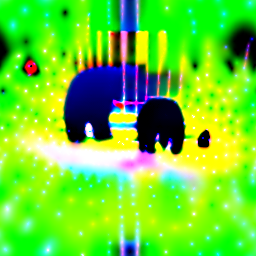} &
        \includegraphics[width=\linewidth]{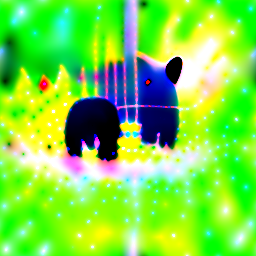} &
        \includegraphics[width=\linewidth]{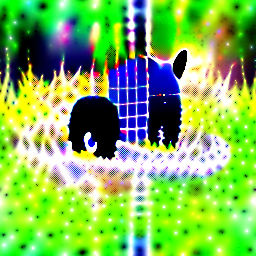} &
        \includegraphics[width=\linewidth]{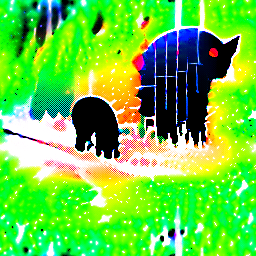} &
        \includegraphics[width=\linewidth]{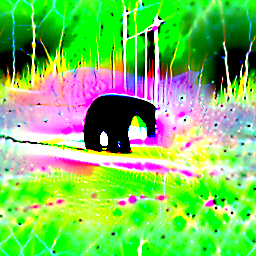} \\

        \raisebox{0.25\totalheight}{\rotatebox[origin=lB]{90}{target=350} } & 
        \includegraphics[width=\linewidth]{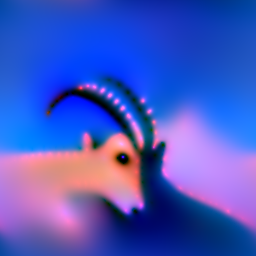} &
        \includegraphics[width=\linewidth]{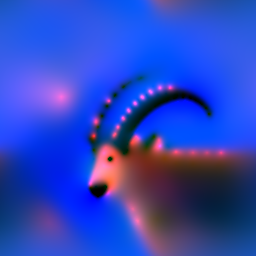} &
        \includegraphics[width=\linewidth]{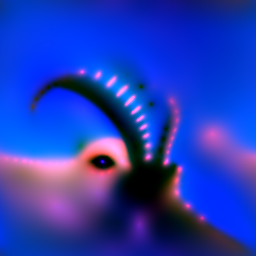} &
        \includegraphics[width=\linewidth]{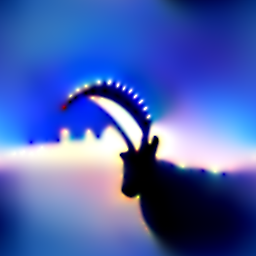} &
        \includegraphics[width=\linewidth]{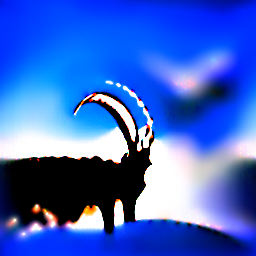} &
        \includegraphics[width=\linewidth]{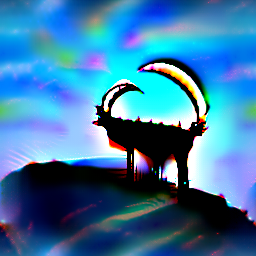} \\

%
%
%
        \raisebox{0.1\totalheight}{\rotatebox[origin=lB]{90}{target=240} } & 
        \includegraphics[width=\linewidth]{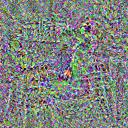} &
        \includegraphics[width=\linewidth]{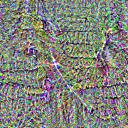} &
        \includegraphics[width=\linewidth]{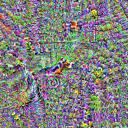} &
        \includegraphics[width=\linewidth]{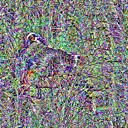} &
        \includegraphics[width=\linewidth]{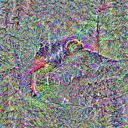} &
        \includegraphics[width=\linewidth]{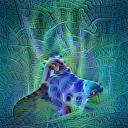} \\

        \raisebox{0.1\totalheight}{\rotatebox[origin=lB]{90}{target=400} } & 
        \includegraphics[width=\linewidth]{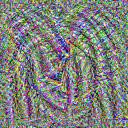} &
        \includegraphics[width=\linewidth]{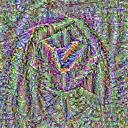} &
        \includegraphics[width=\linewidth]{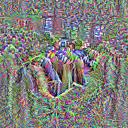} &
        \includegraphics[width=\linewidth]{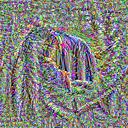} &
        \includegraphics[width=\linewidth]{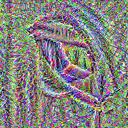} &
        \includegraphics[width=\linewidth]{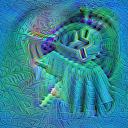} \\

        \raisebox{0.1\totalheight}{\rotatebox[origin=lB]{90}{target=460} } & 
        \includegraphics[width=\linewidth]{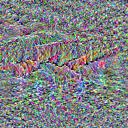} &
        \includegraphics[width=\linewidth]{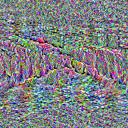} &
        \includegraphics[width=\linewidth]{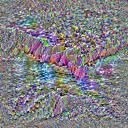} &
        \includegraphics[width=\linewidth]{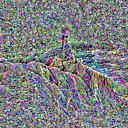} &
        \includegraphics[width=\linewidth]{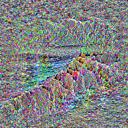} &
        \includegraphics[width=\linewidth]{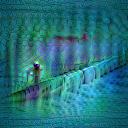} \\

            \end{tabularx}
        \caption{ Effect of using different augmentations on inverting a robustly-trained ResNet-50 (top 3 rows) and a naturally-trained ResNet-50 (bottom 3 rows).}
        \label{fig:app:augablation_natural}
\end{figure}

\subsection{\methodname{} on additional networks}
\begin{figure}[h]
    \centering
    \setlength\tabcolsep{0.5pt}
    \begin{tabularx}{0.97\linewidth}{YYYYYYYY}
            
        \includegraphics[width=\linewidth]{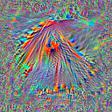} & 
        \includegraphics[width=\linewidth]{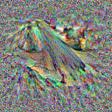} & 
        \includegraphics[width=\linewidth]{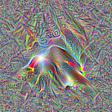} & 
        \includegraphics[width=\linewidth]{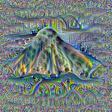} &
        \includegraphics[width=\linewidth]{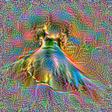} &
        \includegraphics[width=\linewidth]{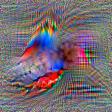} &
        \includegraphics[width=\linewidth]{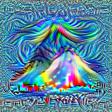} &
        \includegraphics[width=\linewidth]{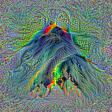} \\ 
        
        AlexNet & DenseNet & GoogLeNet & MobileNet v2 & MobileNet v3-l & MobileNet v3-s & MNasNet 0-5 & MNasNet 1-0 \\

        \includegraphics[width=\linewidth]{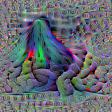} &
        \includegraphics[width=\linewidth]{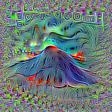} &
        \includegraphics[width=\linewidth]{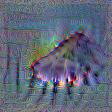} &
        \includegraphics[width=\linewidth]{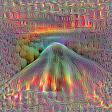} &
        \includegraphics[width=\linewidth]{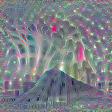} &
        \includegraphics[width=\linewidth]{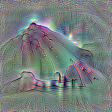} &
        \includegraphics[width=\linewidth]{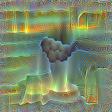} &
        \includegraphics[width=\linewidth]{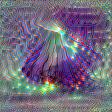} \\
        
        ResNet 18 & ResNet 34 & ResNet 50 & ResNet 101 & ResNet 152 & ResNext 50 & ResNext 101 & WResNet 50 \\

        \includegraphics[width=\linewidth]{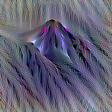} & 
        \includegraphics[width=\linewidth]{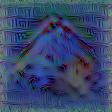} &
        \includegraphics[width=\linewidth]{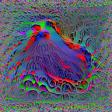} &
        \includegraphics[width=\linewidth]{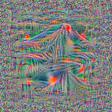} &
        \includegraphics[width=\linewidth]{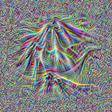} &
        \includegraphics[width=\linewidth]{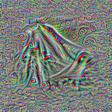} & 
        \includegraphics[width=\linewidth]{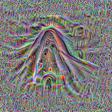} &
        \includegraphics[width=\linewidth]{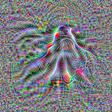} \\
        
        WResNet 101 & ShuffleNet v2-0-5 & ShuffleNet v2-1-0 & SqueezeNet & VGG11-bn & VGG13-bn & VGG16-bn & VGG19-bn \\

        \includegraphics[width=\linewidth]{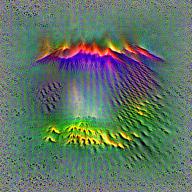} &
        \includegraphics[width=\linewidth]{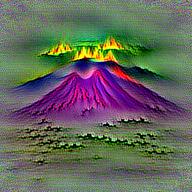} &
        \includegraphics[width=\linewidth]{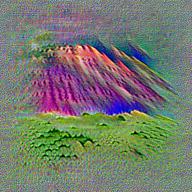} & 
        \includegraphics[width=\linewidth]{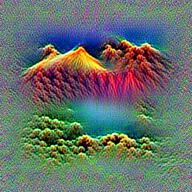} &
        \includegraphics[width=\linewidth]{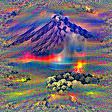} &
        \includegraphics[width=\linewidth]{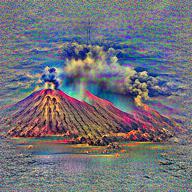} &
        \includegraphics[width=\linewidth]{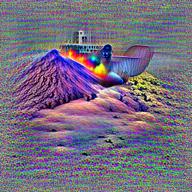} &
        \includegraphics[width=\linewidth]{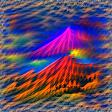} \\
        
        ViT B16 & ViT B32 & ViT L16 & ViT L32 & DeiT p16-224 & DeiT-D p16-384 & Deit p16-384 & DeiT-D-t p16-224 \\

        \includegraphics[width=\linewidth]{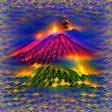} &
        \includegraphics[width=\linewidth]{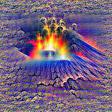} &
        \includegraphics[width=\linewidth]{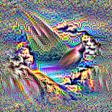} &
        \includegraphics[width=\linewidth]{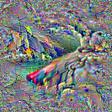} &
        \includegraphics[width=\linewidth]{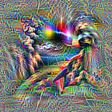} &
        \includegraphics[width=\linewidth]{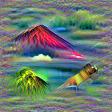} &
        \includegraphics[width=\linewidth]{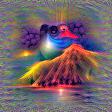} &
        \includegraphics[width=\linewidth]{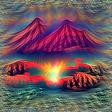} \\

        DeiT-D-s p16-224 & DeiT-D p16-224 & CoaT-m & CoaT-s  & CoaT-t & ConViT & ConViT-s & ConViT-t \\ 
        
        \includegraphics[width=\linewidth]{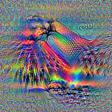} &
        \includegraphics[width=\linewidth]{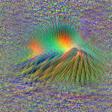} &
        \includegraphics[width=\linewidth]{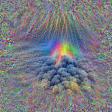} &
        \includegraphics[width=\linewidth]{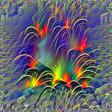} &
        \includegraphics[width=\linewidth]{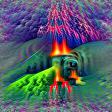} &
        \includegraphics[width=\linewidth]{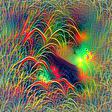} &
        \includegraphics[width=\linewidth]{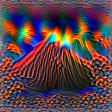} & 
        \includegraphics[width=\linewidth]{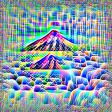} \\
        
        Mixer 24-224 & Mixer b16-224 & Mixer l16-224 & PiT-D b-224 & PiT s-224 & PiT-D s-224 & Pit-D t-224 & ResMLP 12-224 \\ 
        
        \includegraphics[width=\linewidth]{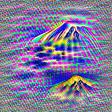} &
        \includegraphics[width=\linewidth]{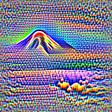} &
        \includegraphics[width=\linewidth]{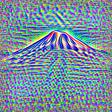} &
        \includegraphics[width=\linewidth]{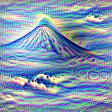} & 
        \includegraphics[width=\linewidth]{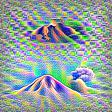} &
        \includegraphics[width=\linewidth]{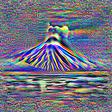} &
        \includegraphics[width=\linewidth]{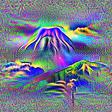} &
        \includegraphics[width=\linewidth]{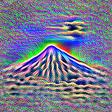} \\
        
        ResMLP-D 12-224 & ResMLP 24-224 & ResMLP-D 24-224 & ResMLP 36-224 & ResMLP-D 36-224 & ResMLP b-24-224 & ResMLP b-24-224-1k & ResMLP-D b-24-224 \\
        
        \includegraphics[width=\linewidth]{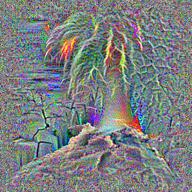} &
        \includegraphics[width=\linewidth]{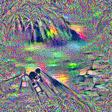} &
        \includegraphics[width=\linewidth]{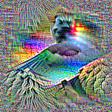} & 
        \includegraphics[width=\linewidth]{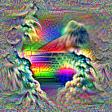} &
        \includegraphics[width=\linewidth]{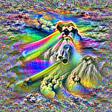} &
        \includegraphics[width=\linewidth]{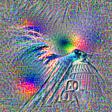} &
        \includegraphics[width=\linewidth]{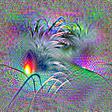} &
        \includegraphics[width=\linewidth]{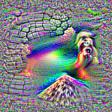} \\
        
        Swin w12-384 & Swin s-w7-224 & Twin pcpvt-b & Twin pcpvt-l & Twin pcpvt-s & Twin svt-b & Twin svt-l & Twin svt-s \\

    \end{tabularx}
    \caption{\text{\methodname} applied to various vision models for the Volcano class.}
    \label{fig:app_various_nets}
\end{figure}

Figure \ref{fig:app_various_nets} shows the results of Plug-In Inversion on various CNN, ViT, and MLP networks, adding to those shown in figure \ref{fig:various_nets}. See section \ref{app:model-library} for model details.

\clearpage
\section{Models}\label{app:model-library}
In our experiments, we use publicly available pre-trained models from various sources. The following tables list the models used from each source, along with references to where they are introduced in the literature.

\setlength\tabcolsep{10.0pt}

\begin{figure}[h]
    \centering
    \begin{tabularx}{\linewidth}{YYY}
        \toprule
        Alias & Name & Paper \\ \hline
        \text{ViT B16} & \text{B\_16\_imagenet1k} & \cite{dosovitskiy2021image} \\
        \text{ViT B32} & \text{B\_32\_imagenet1k} & \cite{dosovitskiy2021image} \\
        \text{ViT B-32} & \text{B\_32\_imagenet1k} & \cite{dosovitskiy2021image} \\
        \text{ViT L16} & \text{L\_16\_imagenet1k} & \cite{dosovitskiy2021image} \\
        \text{ViT L32} & \text{L\_32\_imagenet1k} & \cite{dosovitskiy2021image} \\
      
        \bottomrule
    \end{tabularx}
    \caption{ 
        Pre-trained models used from : \href{https://github.com/lukemelas/PyTorch-Pretrained-ViT}{https://github.com/lukemelas/PyTorch-Pretrained-ViT}. 
    }
    \label{table:app:model:vit}
\end{figure}

\begin{figure}[h]
    \centering
    \begin{tabularx}{\linewidth}{YYY}
        \toprule
        Alias & Name  & Paper \\ \hline
 
        \text{DeiT p16-224} & \text{deit\_base\_patch16\_224} &   \cite{touvron2021training}  \\
        \text{DeiT P16 224} & \text{deit\_base\_patch16\_224} &   \cite{touvron2021training}  \\
        \text{Deit-D p16-384} & \text{deit\_base\_distilled\_patch16\_384} &    \cite{touvron2021training}  \\
        \text{Deit Dist P16 384} & \text{deit\_base\_distilled\_patch16\_384} &    \cite{touvron2021training}  \\
        \text{deit p16-384} & \text{deit\_base\_patch16\_384} &   \cite{touvron2021training}  \\
        \text{Deit-D-t p16-224} & \text{deit\_tiny\_distilled\_patch16\_224} &    \cite{touvron2021training}  \\
        \text{Deit-D-s p16-224} & \text{deit\_small\_distilled\_patch16\_224} &   \cite{touvron2021training}  \\
        \text{Deit-D p16-224} & \text{deit\_base\_distilled\_patch16\_224} &   \cite{touvron2021training}  \\
        \bottomrule
    \end{tabularx}
    \caption{ Pre-trained models from \cite{pmlrv139touvron21a} . }
    \label{table:app:model:deit}
\end{figure}

\begin{figure}[h]
    \centering
    \begin{tabularx}{\linewidth}{YYY}
        \toprule
        Alias & Name & Paper \\ \hline
            
        \text{AlexNet} & \text{alexnet}  &  \cite{krizhevsky2012imagenet} \\
        \text{DenseNet} & \text{densenet121}  &  \cite{huang2017densely} \\
        \text{GoogLeNet} & \text{googlenet}  &  \cite{szegedy2015going} \\
        \text{MobileNet v2} & \text{mobilenet\_v2}  &  \cite{sandler2018mobilenetv2} \\
        \text{MobileNet-v2} & \text{mobilenet\_v2}  &  \cite{sandler2018mobilenetv2} \\
        \text{MobileNet v3-l} & \text{mobilenet\_v3\_large}  &  \cite{howard2019searching} \\
        \text{MobileNet v3-s} & \text{mobilenet\_v3\_small}  &  \cite{howard2019searching} \\
        \text{MNasNet 0-5} & \text{mnasnet0\_5}  &  \cite{tan2019mnasnet} \\
        \text{MNasNet 1-0} & \text{mnasnet1\_0}  &  \cite{tan2019mnasnet} \\
        \text{ResNet 18} & \text{resnet18}  &  \cite{he2016deep} \\
        \text{ResNet-18} & \text{resnet18}  &  \cite{he2016deep} \\
        \text{ResNet 34} & \text{resnet34}  &  \cite{he2016deep} \\
        \text{ResNet 50} & \text{resnet50}  &  \cite{he2016deep} \\
        \text{ResNet 101} & \text{resnet101}  & 
        \cite{he2016deep} \\
        \text{ResNet-101} & \text{resnet101}  & \cite{he2016deep} \\
        \text{ResNet 152} & \text{resnet152}  &  \cite{he2016deep} \\
        \text{ResNext 50} & \text{resnext50\_32x4d}  &  \cite{xie2017aggregated} \\
        \text{ResNext 101} & \text{resnext101\_32x8d}  &  \cite{xie2017aggregated} \\
        \text{WResNet 50} & \text{wide\_resnet50\_2}  &  \cite{zagoruyko2016wide} \\
        \text{WResNet 101} & \text{wide\_resnet101\_2}  &
        \cite{zagoruyko2016wide} \\
        \text{W-ResNet-101-2} & \text{wide\_resnet101\_2}  &\cite{zagoruyko2016wide} \\
        \text{ShuffleNet v2-0-5} & \text{shufflenet\_v2\_x0\_5}  &   \cite{ma2018shufflenet} \\
        \text{ShuffleNet v2-1-0} & \text{shufflenet\_v2\_x1\_0}  &   \cite{ma2018shufflenet} \\
        \text{ShuffleNet v2} & \text{shufflenet\_v2\_x1\_0}  &   \cite{ma2018shufflenet} \\
        \text{SqueezeNet} & \text{squeezenet1\_0}  &  \cite{iandola2016squeezenet} \\
        \text{VGG11-bn} & \text{vgg11\_bn}  &   \cite{simonyan2014very} \\
        \text{VGG13-bn} & \text{vgg13\_bn}  &   \cite{simonyan2014very}\\
        \text{VGG16-bn} & \text{vgg16\_bn}  &   \cite{simonyan2014very} \\
        \text{VGG19-bn} & \text{vgg19\_bn}  &   \cite{simonyan2014very} \\
 
        \bottomrule
    \end{tabularx}
    \caption{ Pre-trained models from TorchVision: \href{https://github.com/pytorch/vision}{https://github.com/pytorch/vision}. }
    \label{table:app:model:torchvision}
\end{figure}

\begin{figure}[h]
    \centering
    \begin{tabularx}{\linewidth}{YYY}
        \toprule
        Alias & Name &  Paper \\ \hline
        
        \text{CoaT-m} & \text{coat\_lite\_mini} &  \cite{xu2021co} \\
        \text{CoaT-s} & \text{coat\_lite\_small} &  \cite{xu2021co} \\
        \text{CoaT-t} & \text{coat\_lite\_tiny} &  \cite{xu2021co} \\
        \text{ConViT} & \text{convit\_base} &   \cite{d2021convit} \\
        \text{ConViT-s} & \text{convit\_small} &   \cite{d2021convit} \\
        \text{ConViT-t} & \text{convit\_tiny} &   \cite{d2021convit} \\
        \text{ConViT tiny} & \text{convit\_tiny} &   \cite{d2021convit} \\
        \text{Mixer 24-224} & \text{mixer\_24\_224} &  \cite{tolstikhin2021mlp} \\
        \text{Mixer b16-224} & \text{mixer\_b16\_224} &  \cite{tolstikhin2021mlp} \\
        \text{Mixer b16 224} & \text{mixer\_b16\_224} &  \cite{tolstikhin2021mlp} \\
        \text{Mixer b16-224-mill} & \text{mixer\_b16\_224\_miil} &  \cite{tolstikhin2021mlp} \\
        \text{Mixer l16-224} & \text{mixer\_l16\_224} &  \cite{tolstikhin2021mlp} \\
        \text{PiT-D b-224} & \text{pit\_b\_distilled\_224} &  \cite{heo2021rethinking} \\
        \text{PiT Dist 224} & \text{pit\_b\_distilled\_224} &  \cite{heo2021rethinking} \\
        \text{PiT s-224} & \text{pit\_s\_224} &  \cite{heo2021rethinking} \\
        \text{PiT-D s-224} & \text{pit\_s\_distilled\_224} &  \cite{heo2021rethinking} \\
        \text{PiT-D t-224} & \text{pit\_ti\_distilled\_224} &  \cite{heo2021rethinking} \\
        \text{ResMLP 12-224} & \text{resmlp\_12\_224} &  \cite{touvron2021resmlp}\\
        \text{ResMLP-D 12-224} & \text{resmlp\_12\_distilled\_224} &  \cite{touvron2021resmlp} \\
        \text{ResMLP 24-224} & \text{resmlp\_24\_224} &  \cite{touvron2021resmlp} \\
        \text{ResMLP-D 24-224} & \text{resmlp\_24\_distilled\_224} &  \cite{touvron2021resmlp} \\
        \text{ResMLP 36-224} & \text{resmlp\_36\_224} &  \cite{touvron2021resmlp} \\
        \text{ResMLP-D 36-224} & \text{resmlp\_36\_distilled\_224} &  \cite{touvron2021resmlp} \\
        \text{ResMLP 36 Dist} & \text{resmlp\_36\_distilled\_224} &  \cite{touvron2021resmlp} \\
        \text{ResMLP b-24-224} & \text{resmlp\_big\_24\_224} &  \cite{touvron2021resmlp} \\
        \text{ResMLP b-24-224-1k} & \text{resmlp\_big\_24\_224\_in22ft1k} &  \cite{touvron2021resmlp} \\
        \text{ResMLP-D b-24-224} & \text{resmlp\_big\_24\_distilled\_224} &  \cite{touvron2021resmlp} \\
        \text{Swin w7-224} & \text{swin\_base\_patch4\_window7\_224} &  \cite{liu2021swin}\\
        \text{Swin l-w7-224} & \text{swin\_large\_patch4\_window7\_224} &  \cite{liu2021swin}\\
        \text{Swin l-w12-384} & \text{swin\_large\_patch4\_window12\_384} &  \cite{liu2021swin}\\
        \text{Swin w12-384} & \text{swin\_base\_patch4\_window12\_384} &  \cite{liu2021swin}\\
        \text{Swin P4 W12} &
        \text{swin\_base\_patch4\_window12\_384} &  \cite{liu2021swin}\\
        \text{Swin s-w7-224} & \text{swin\_small\_patch4\_window7\_224} &  \cite{liu2021swin} \\
        \text{Swin t-w7-224} & \text{swin\_tiny\_patch4\_window7\_224} &  \cite{liu2021swin} \\
        \text{Twin pcpvt-b} & \text{twins\_pcpvt\_base} &  \cite{chu2021twins} \\
        \text{Twin PCPVT} & \text{twins\_pcpvt\_base} &  \cite{chu2021twins} \\
        \text{Twins pcpvt-l} & \text{twins\_pcpvt\_large} &  \cite{chu2021twins}\\
        \text{Twins pcpvt-s} & \text{twins\_pcpvt\_small} &  \cite{chu2021twins} \\
        \text{Twins svt-b} & \text{twins\_svt\_base} &  \cite{chu2021twins} \\
        \text{Twins svt-l} & \text{twins\_svt\_large} &  \cite{chu2021twins} \\
        \text{Twins svt-s} & \text{twins\_svt\_small} &  \cite{chu2021twins} \\
        \bottomrule
    \end{tabularx}
    \caption{Pre-trained models used from: \cite{rw2019timm} }
    \label{table:app:weightman.}
\end{figure}

\clearpage

\section{Additional experimental setting}\label{app:robust-setting}
\subsection{Robust models}
We use a robust RestNet-50 \citep{he2016deep} model free-trained \citep{shafahi2019adversarial} on the ImageNet dataset \citep{deng2009imagenet}. The setting we use for inverting robust models is very similar to that of \methodname{} explained in section \ref{exp-setup} except for some differences. Throughout the paper, we use centering for robust models unless otherwise is mentioned (like when we are examining the effect of zoom and centering themselves). We use 0.0005 to scale total variation in the loss function. Also, we do not apply the data normalization layer before feeding the input to the network. In \methodname{} experiment setting, we apply a random ColorShift at each optimization step to each element in the ensemble. In the robust setting, we do not update the ColorShift variables $\mu$, and $\sigma$ for a fixed patch size, and we update these variables for the ensemble when we use a new patch size. Although using \augname{} would alleviate the need for using TV regularization as discussed in section \ref{plugin:color-jitter}, and illustrated in figure \ref{fig:tv}, we retain the TV penalty in our robust setting to make this setting more similar to that of previous inversion methods and to emphasize that it is a toy example for our ablation studies.

\section{Every Class of ImageNet Dataset Inverted}

\begin{figure}[h]
    \centering
    \includegraphics[width=\linewidth]{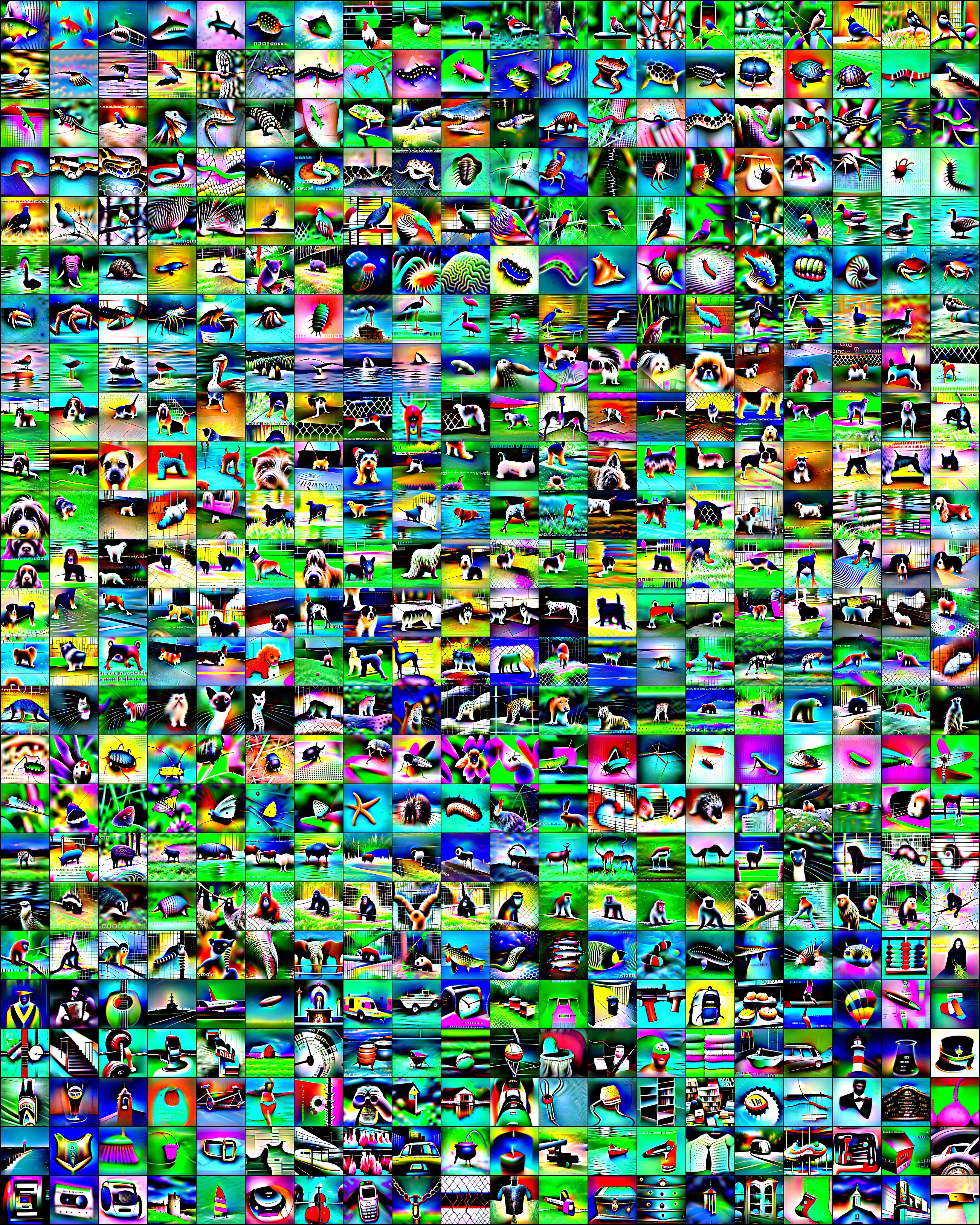}
    \caption{Inversion of first 500 classes of ImageNet for the Robust Model.}
    \label{fig:app:all_classes1}
\end{figure}

\begin{figure}[h]
    \centering
    \includegraphics[width=\linewidth]{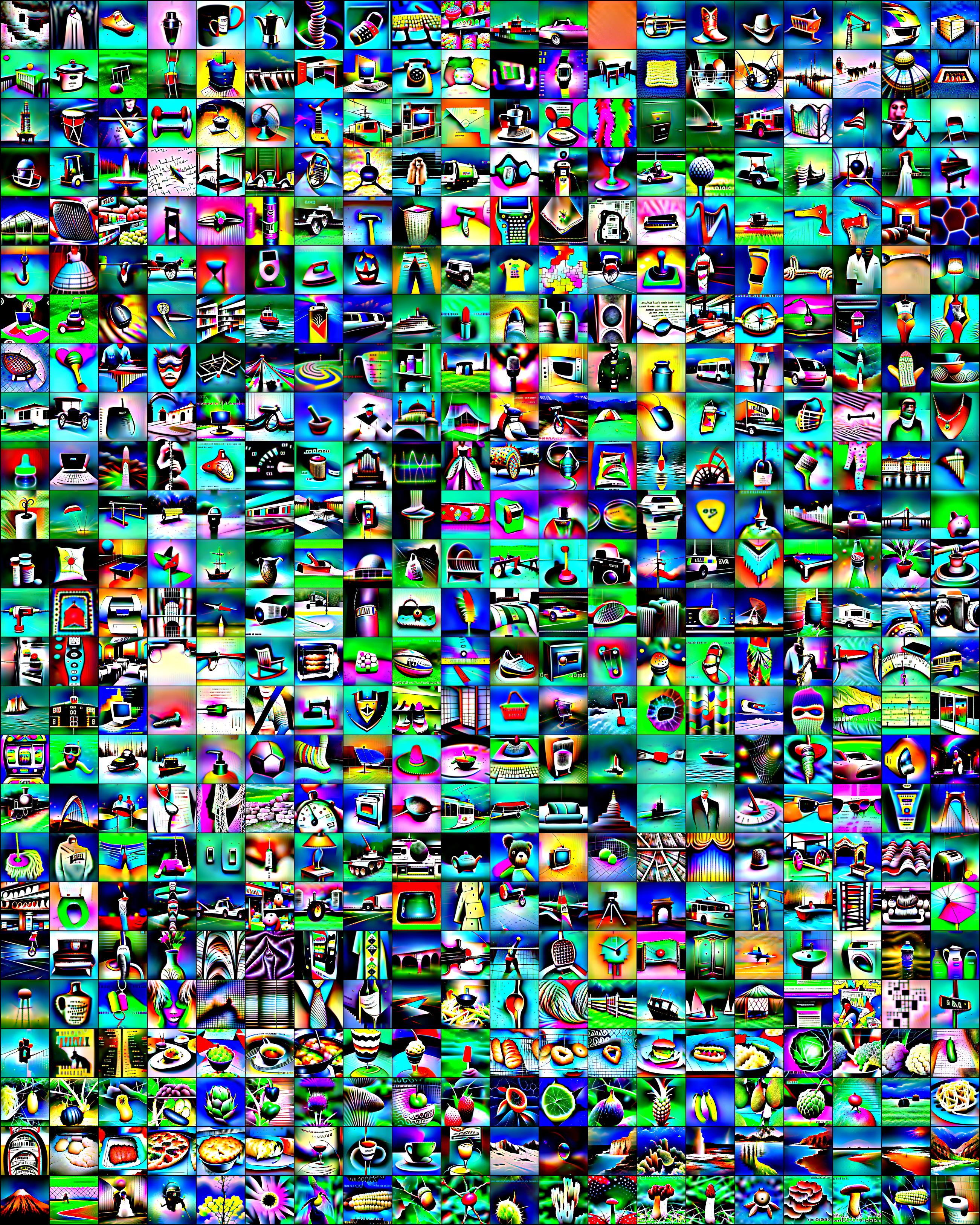}
    \caption{Inversion of second 500 classes of ImageNet for the Robust Model.}
    \label{fig:app:all_classes2}
\end{figure}

\section{Optimization algorithm}\label{sec:algorithm}

\begin{algorithm}
\caption{Optimization procedure for Plug-In Inversion}
\label{alg:pii}
\SetKwInOut{Input}{Input}
\Input{Model $f$, class $y$, final resolution $R$, ColorShift parameters $\alpha, \beta$, `ensemble' size $e$, randomly initialized $\mathbf{x} \in \mathcal{I}^{3 \times R/8 \times R/8}$}
\For{$s=1, \dots, 7$}{
Upsample $\mathbf{x}$ to resolution $\frac{(2s+1)R}{16} \times \frac{(2s+1)R}{16}$ \\
Pad $\mathbf{x}$ with random noise to resolution $\frac{(s+1)R}{8} \times \frac{(s+1)R}{8}$ \\
\For{$i = 1, \dots, 400$}{
$\mathbf{x}' = \text{Jitter}(\mathbf{x})$ \\
\For{$n = 1, \dots, e$}{
Draw $\mu \sim U(-\alpha, \alpha)^3$, $\sigma \sim \exp(U(-\beta, \beta))^3$ \\
$\mathbf{x}_n = \text{ColorShift}_{\mu, \sigma}(\mathbf{x}')$
}
$\displaystyle \mathcal{L} = \frac{1}{e} \sum_{n=1}^e NLL(f(\mathbf{x_n}), y)$ \\
$\mathbf{x} \leftarrow \text{Adam}_i(\mathbf{x}, \nabla_\mathbf{x} \mathcal{L})$
}
\Return $\mathbf{x}$
}
\end{algorithm}

\section{Additional baseline comparisons}\label{sec:baselines}


\def \varfiglen{0.19\linewidth}
\begin{figure}[h]
    \centering
        \centering
        \setlength\tabcolsep{1.5pt}
        \begin{tabularx}{\linewidth}{ccccc}
            MobileNet-v2 & ResNet-18 & VGG16-bn & W- ResNet-101-2 & ShuffleNet-v2\\
             
            \includegraphics[width=\varfiglen]{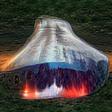} & 
            \includegraphics[width=\varfiglen]{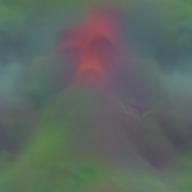} &
            \includegraphics[width=\varfiglen]{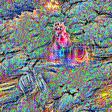} & 
            \includegraphics[width=\varfiglen]{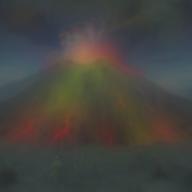} &
            \includegraphics[width=\varfiglen]{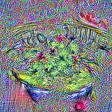} \\
            
            ResNet-101 & ViT B-32 & DeiT P16 224 & Deit Dist P16 384 & ConViT tiny \\ 
            
            \includegraphics[width=\varfiglen]{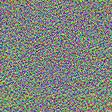} &
            \includegraphics[width=\varfiglen]{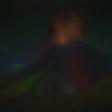} & 
            \includegraphics[width=\varfiglen]{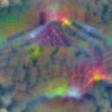} &
            \includegraphics[width=\varfiglen]{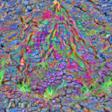} &
            \includegraphics[width=\varfiglen]{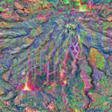} \\
            
            Mixer b16 224 & PiT Dist 224 & ResMLP 36 Dist & Swin P4 W12 & Twin PCPVT \\
            
        \end{tabularx}
        \caption{ Images inverted from the Volcano class for
        various Convolutional, Transformer, and MLP-based networks using DeepInversion (CNN models) / DeepDream (non-CNN models). Cross-reference figure \ref{fig:various_nets}.}
        \label{fig:various_nets_di}
\end{figure}

\def \varvarnetvarclass{0.13\linewidth}
\begin{figure}[h]
    \centering
        \centering
        \setlength\tabcolsep{1.5pt}
        \begin{tabularx}{0.9\linewidth}{ccccccc}
            
            \multirow{2}{*}{} & \multirow{2}{*}{Barn} & Garbage & \multirow{2}{*}{Goblet} & Ocean & CRT & \multirow{2}{*}{Warplane}\\ 
              &  & Truck &  & Liner & Screen &  \\ 

\raisebox{0.05\totalheight}{\rotatebox[origin=lB]{90}{ResNet-101}} &
\includegraphics[width=\varvarnetvarclass]{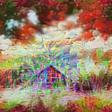}  &
\includegraphics[width=\varvarnetvarclass]{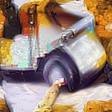}  &
\includegraphics[width=\varvarnetvarclass]{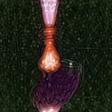}  &
\includegraphics[width=\varvarnetvarclass]{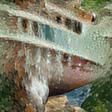}  &
\includegraphics[width=\varvarnetvarclass]{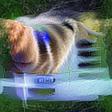}  &
\includegraphics[width=\varvarnetvarclass]{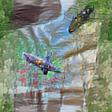}  \\

\raisebox{0.2\totalheight}{\rotatebox[origin=lB]{90}{ViT B-32}} &
\includegraphics[width=\varvarnetvarclass]{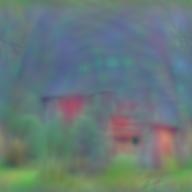}  &
\includegraphics[width=\varvarnetvarclass]{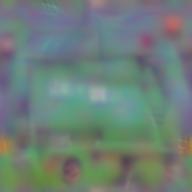}  &
\includegraphics[width=\varvarnetvarclass]{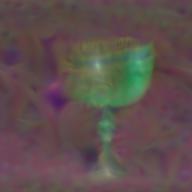}  &
\includegraphics[width=\varvarnetvarclass]{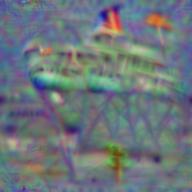}  &
\includegraphics[width=\varvarnetvarclass]{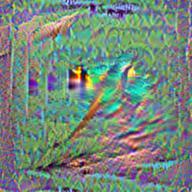}  &
\includegraphics[width=\varvarnetvarclass]{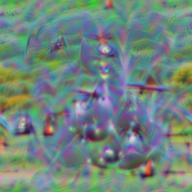}  \\

\raisebox{0.2\totalheight}{\rotatebox[origin=lB]{90}{DeiT Dist}} &
\includegraphics[width=\varvarnetvarclass]{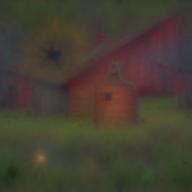}  &
\includegraphics[width=\varvarnetvarclass]{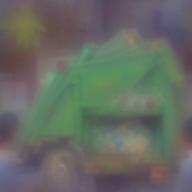}  &
\includegraphics[width=\varvarnetvarclass]{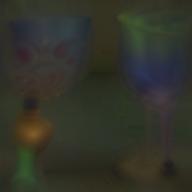}  &
\includegraphics[width=\varvarnetvarclass]{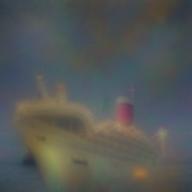}  &
\includegraphics[width=\varvarnetvarclass]{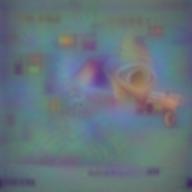}  &
\includegraphics[width=\varvarnetvarclass]{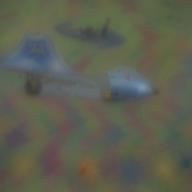}  \\

\raisebox{0.05\totalheight}{\rotatebox[origin=lB]{90}{ResMLP 36}} &
\includegraphics[width=\varvarnetvarclass]{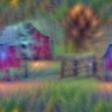}  &
\includegraphics[width=\varvarnetvarclass]{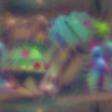}  &
\includegraphics[width=\varvarnetvarclass]{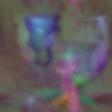}  &
\includegraphics[width=\varvarnetvarclass]{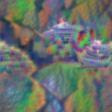}  &
\includegraphics[width=\varvarnetvarclass]{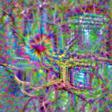}  &
\includegraphics[width=\varvarnetvarclass]{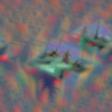}  \\

            \end{tabularx}
        \caption{Inverting different model and class combinations for different classes using DeepInversion (top row) / DeepDream (other rows). Cross-reference figure \ref{fig:various_classes}.}
        \label{fig:various_classes_di}
\end{figure}

\pagebreak

\section{Quantitative Results}\label{sec:quant}

\begin{figure}[h]
    \centering
    \subfloat[]{{\includegraphics[width=\columnwidth]{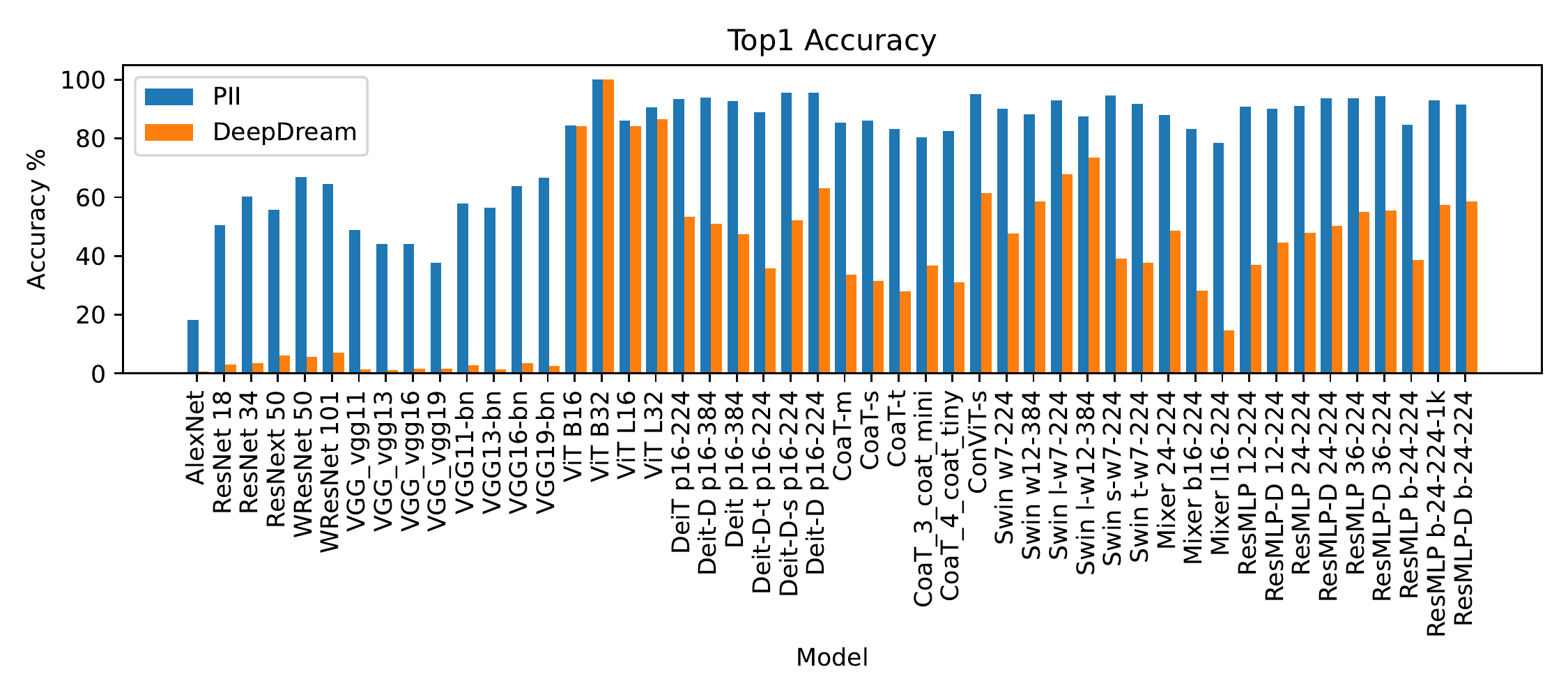} }}
    \qquad
    \subfloat[]{{\includegraphics[width=\columnwidth]{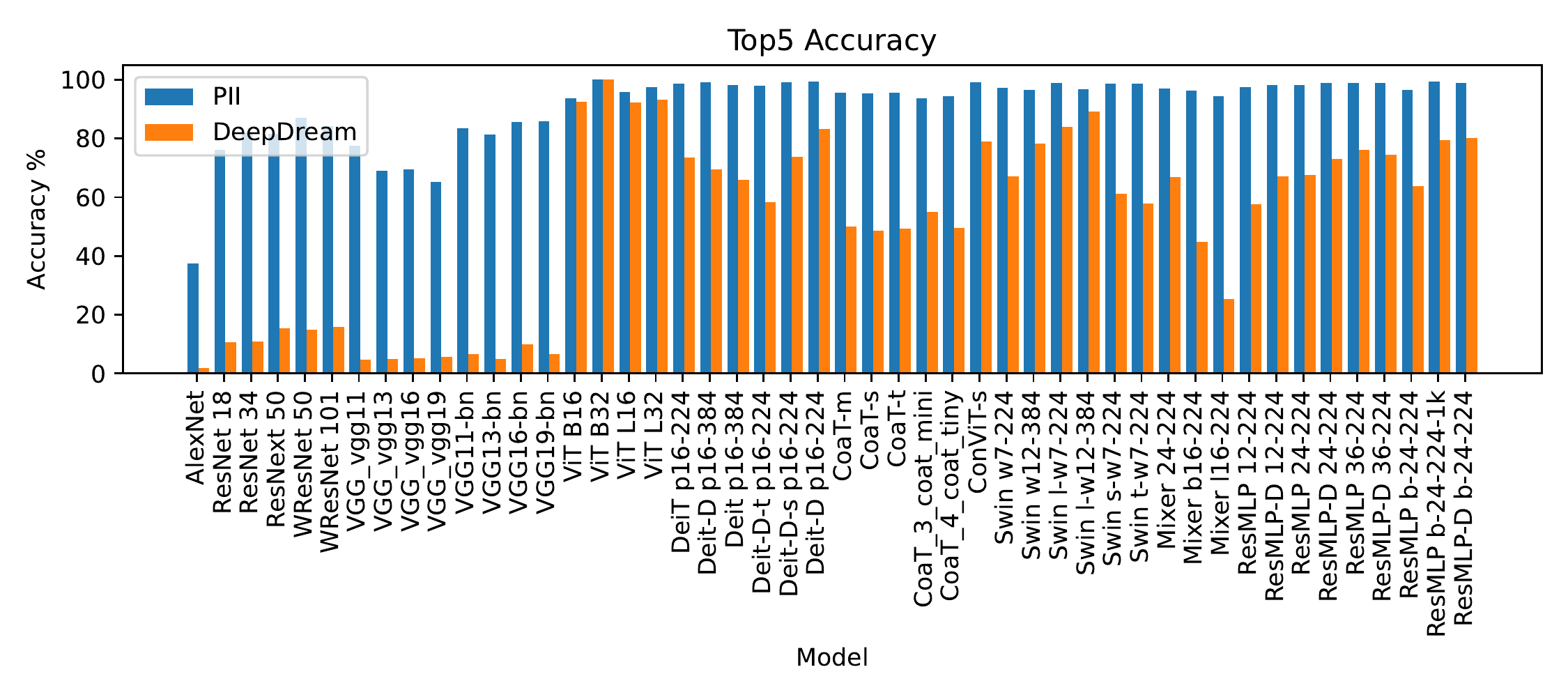} }}
    \caption{Top-1 (a) and top-5 (b) classification accuracy of various CNN, ViT, and MLP models evaluated on images generated from ViT B-32 using PII and DeepDream.}
    \label{fig:vit-classification}
\end{figure}

To quantitatively evaluate our method, we invert a pre-trained ViT model to produce one image per class using \methodname{}, and do the same using DeepDream (i.e., DeepInversion minus feature regularization, which is not available for this model). We then use a variety of pre-trained CNN, ViT, and MLP models to classify these images. We find that every model achieves strictly higher top-1 and top-5 accuracy on the \methodname{}-generated image set (excepting the `teacher' model, which perfectly classifies both). We compile these results in figure \ref{fig:vit-classification}. Additionally, we compute the Inception score \citep{salimans2016improved} for both sets of images, which also favors \methodname{} over DeepDream, with scores of $28.17 \pm 7.21$ and $2.72 \pm 0.23$, respectively.

We also perform the same evaluation for images generated from a pre-trained ResMLP model. These results are more mixed; DeepDream images are classified much better by a small number of models, but the majority of models classify \methodname{} images better, and the average accuracy across models is approximately equal for both methods. Inception score, however, once again clearly favors \methodname{} over DeepDream, with scores of $6.79 \pm 2.18$ and $3.27 \pm 0.47$, respectively.

\begin{figure}[h]
    \centering
    \subfloat[]{{\includegraphics[width=\columnwidth]{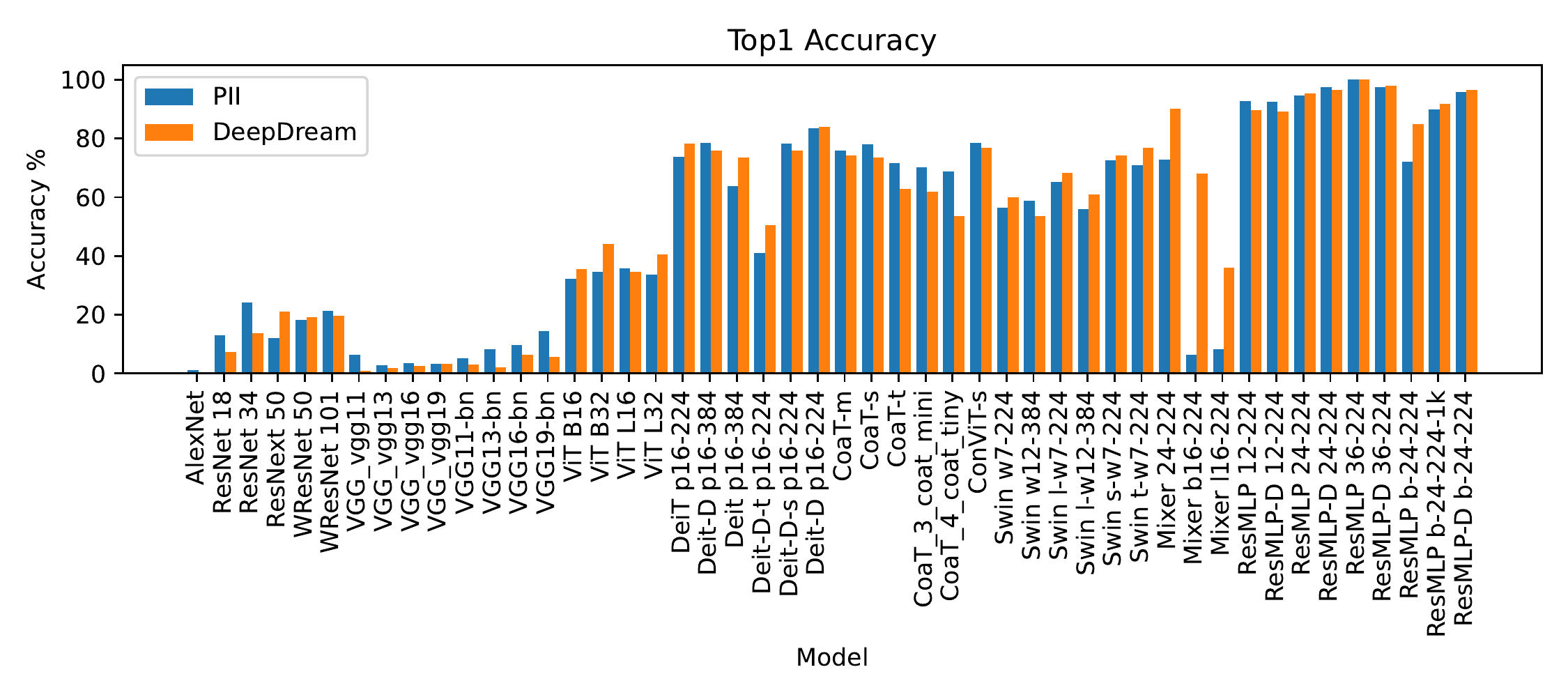} }}
    \qquad
    \subfloat[]{{\includegraphics[width=\columnwidth]{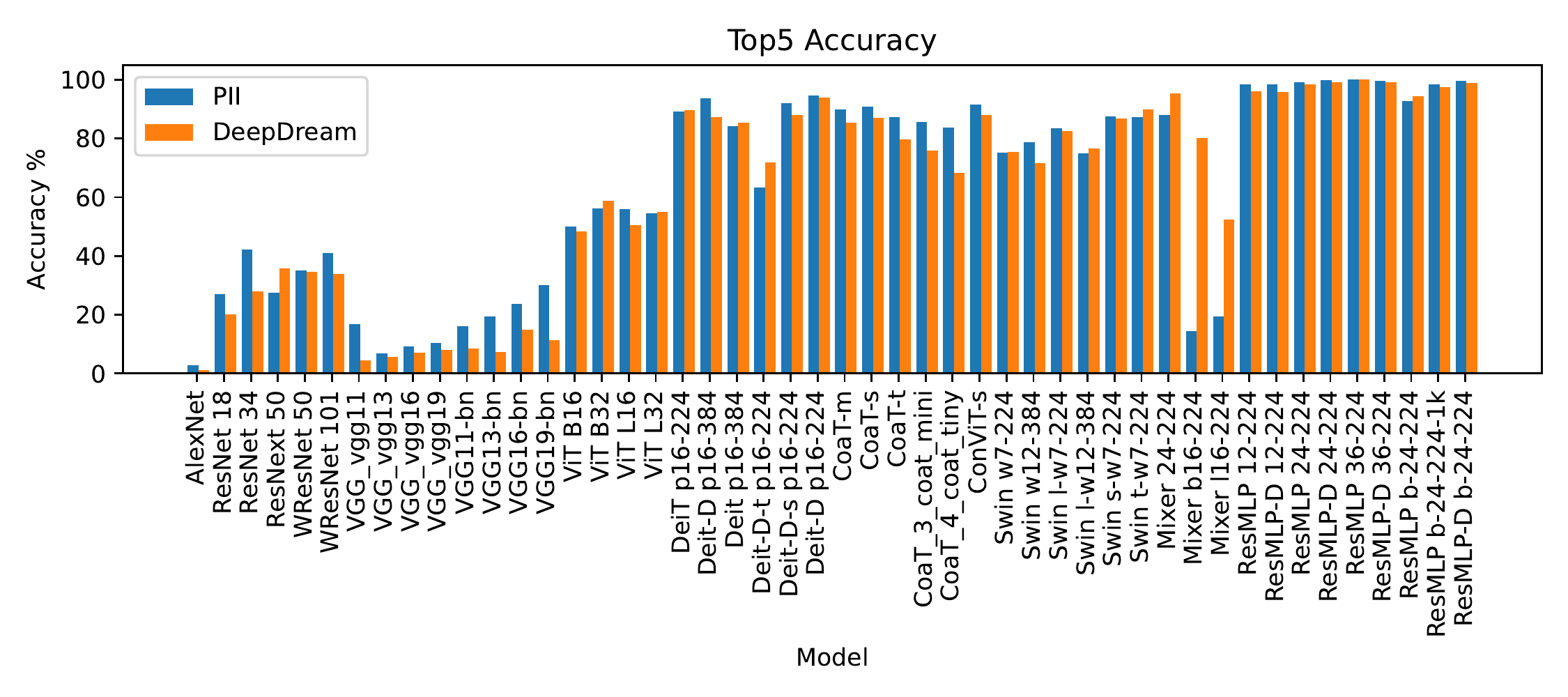} }}
    \caption{Top-1 (a) and top-5 (b) classification accuracy of various CNN, ViT, and MLP models evaluated on images generated from ResMLP 36-224 using PII and DeepDream.}
    \label{fig:mlp-classification}
\end{figure}

\end{document}